\documentclass{fairmeta}
\usepackage{amsmath}
\usepackage{enumerate} 
\usepackage{bm}
\usepackage{algorithm}
\usepackage{floatflt}
\usepackage{algpseudocode}
\usepackage{amsfonts}
\usepackage{amsthm}
\usepackage{newtxtt}
\usepackage{algorithm}
\usepackage{colortbl} 
\usepackage{cleveref}
\usepackage{diagbox} 
\usepackage[utf8]{inputenc}
\usepackage{textgreek}
\usepackage{colortbl}
\usepackage{nicematrix}
\usepackage{makecell}
\usepackage{float}
\usepackage{arydshln}
\usepackage[frozencache,cachedir=.]{minted}
\usepackage{caption}
\usepackage{subcaption}
\usepackage[table]{xcolor}
\usepackage{tcolorbox}
\usepackage{amssymb}
\usepackage{xspace}
\usepackage{wrapfig}
\usepackage{adjustbox}
\usepackage{tabularx}
\usepackage{booktabs}
\usepackage{mathtools}
\usepackage{wrapfig}
\usepackage{amssymb}
\usepackage{graphicx}
\usepackage{pifont}

\newcolumntype{C}{>{\centering\arraybackslash}X}

\usepackage{silence}
\makeatletter
\patchcmd{\wrong@fontshape}{\@gobbletwo}{}{}{}
\makeatother
\definecolor{upColor}{RGB}{17,138,21}
\definecolor{downColor}{RGB}{174,36,67}

\newtheorem{theorem}{Theorem}[]

\newtheorem{remark1}[theorem]{Remark}

\title{Mixture-of-Thought-Tokens: Unifying Perception and Reasoning for Free-form Multimodal Grounding}
\author[1,2,\textsuperscript{$\dagger$}]{Tianyi Gao}
\author[2,\textsuperscript{$\dagger$}]{Han Fang}
\author[2,4]{Tianyi Ding}
\author[2,5]{Hao Li}
\author[2]{Xin Wei}
\author[2]{Hongbo Sun}
\author[2]{Xiaodong Dong}
\author[2]{Ye Yuan}
\author[3]{Jinglin Xu}
\author[4]{Kongming Liang}
\author[2,*]{Hao Sun}
\author[1,*]{Jingmin Xin}
\affiliation[1]{State Key Laboratory of Human-Machine Hybrid Augmented Intelligence, Institute of Artificial Intelligence and Robotics, Xi'an Jiaotong University}
\affiliation[2]{Xingchen AGI Lab, China Telecom Artificial Intelligence Technology (Beijing) Co., Ltd}
\affiliation[3]{University of Science and Technology Beijing}
\affiliation[4]{Beijing University of Posts and Telecommunications}
\affiliation[5]{Shanghai Jiao Tong University}
\metadata[Links]{ \href{https://tg0110.github.io/prbench.github.io/}{Project Page} $|$ \href{https://github.com/TG0110/Motto}{GitHub} } 
\date{July 28, 2026}

\begin{document}

\abstract{
Multimodal Large Language Models have made great progress in grounding tasks, yet existing methods still struggle to unify precise localization and complex reasoning.
For one thing, text-based methods rely on  coordinates or index prediction, severely limiting the perceptual capabilities of the model for dense visual objects.
Meanwhile, latent token-based methods  employ special tokens without inherent spatial references and use a decoding mechanism that lacks thinking steps, weakening high-level reasoning capabilities. Consequently, developing a unified framework that excels in both perception and reasoning remains challenging.
To address this, we propose Mixture-of-Thought-Tokens (Motto), a new free-form multimodal grounding method that bridges the perception-reasoning gap, enabling MLLMs to empower diverse, arbitrary grounding queries.
Specifically, we introduce Spatially-Grounded Thought Tokenization to explicitly align special tokens with spatial locations for clear spatial correspondence and visual interpretability. 
We further design a Context-Adaptive Chain-of-Tokens that dynamically switch grounding modes within an interleaved reasoning chain, achieving robust grounding across tasks of varying complexity. 
In addition, we construct PR-Bench, a new referring expression comprehension benchmark to evaluate the perception-reasoning gap. Extensive experiments demonstrate that Motto achieves state-of-the-art performance across diverse free-form  grounding tasks.
}

\maketitle

\begingroup
\renewcommand{\thefootnote}{}
\footnotetext{%
  \rmfamily\bfseries
  \textsuperscript{$\dagger$}Equal Contributions
  \quad
  \textsuperscript{*}Corresponding Authors
}
\addtocounter{footnote}{-1}
\endgroup

\section{Introduction}

\begin{figure}[t]
  \includegraphics[width=\textwidth]{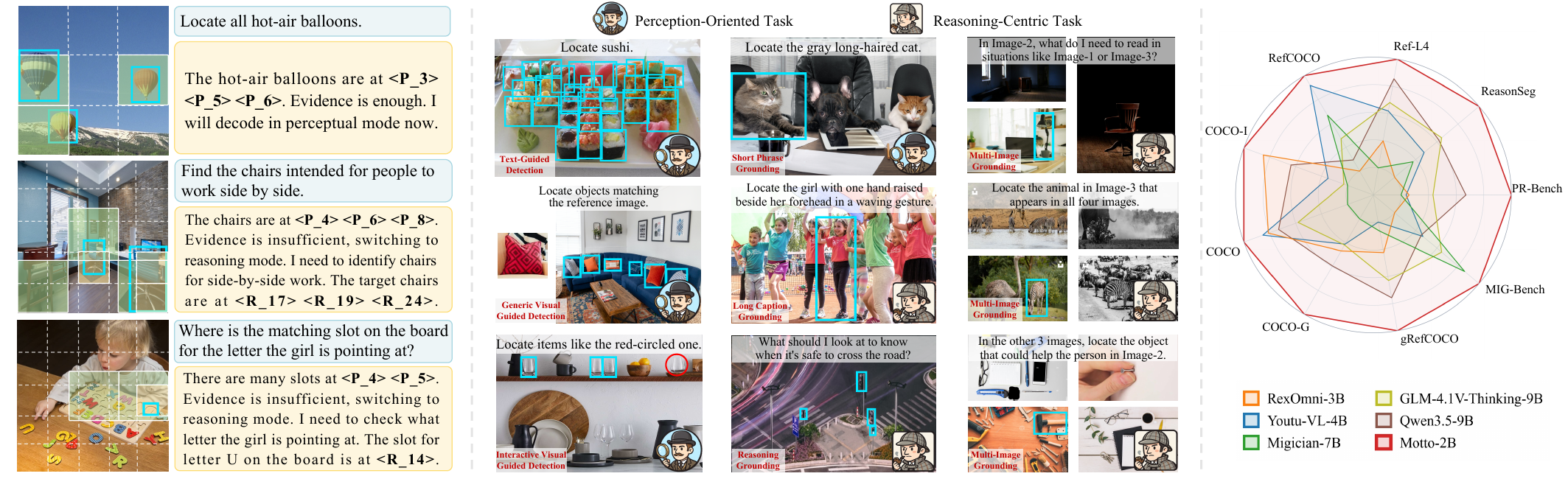}
  \caption{Mixture-of-Thought-Tokens (\textbf{Motto}) tackles a wide range of  grounding tasks by integrating Spatially-Grounded Thought Tokenization and Context-Adaptive Chain-of-Tokens, thereby unifying perception and reasoning capabilities. }
  \label{fig:teaser}
\end{figure}

Fine-grained visual perception and reasoning, which aims to associate semantic queries with  image regions, serves as a challenging task in computer vision~\cite{yan2025task, wu2024visual, 10888636}. 
Previous efforts primarily  focus on open-vocabulary detection~\cite{zareian2021open, liu2024grounding} via class prompts, adopting dual encoders with separate features and cross-modal interaction. 
However, such paradigms exhibit limitations in handling intricate referring expressions due to insufficient language comprehension. 
More critically, their reliance on rigid pairwise cross-modal fusion severely limits generalization to  arbitrary query combinations.

Recent advances in Multimodal Large Language Models (MLLMs) have driven progress in free-form multimodal grounding via unified encoding and pre-trained knowledge~\cite{singh2025gpt5_system_card,  yang2025qwen3}.
For example, UFO~\cite{tangufo} unifies detection and segmentation for fine-grained perception.
Rex-Omni~\cite{jiang2025detect} handles diverse prompts via relative coordinates.
Migician~\cite{li2025migician} enhances reasoning via chain-of-thought mechanisms. 
Despite these, existing  methods still suffer from a critical limitation: they are inherently designed to excel either in low-level visual perception or high-level semantic reasoning, failing to unify these two capabilities within a single framework.
As illustrated in Figure~\ref{fig:teaser}, this performance gap stems from the divergent task priorities between perception and reasoning. 
Perception-oriented tasks (e.g., text/visual-guided detection{~\cite{lin2014microsoft, tang2024lovd}) prioritize capturing extensive visual cues to achieve accurate localization of multiple targets. In contrast, reasoning-centric tasks (e.g., referring expression comprehension~\cite{mao2016generation, xiao2024hivg}, reasoning grounding~\cite{lai2024lisa}, and multi-image grounding~\cite{li2025migician}) require semantic understanding to resolve visual ambiguities for identifying the specific target. Consequently, developing a unified framework that addresses both perception and reasoning for free-form query combinations remains a challenge.

Current MLLM-based methods can be generally categorized into three strategies: (i) direct coordinate prediction~\cite{liu2025visual, zhan2024griffon}, (ii) detect-and-retrieval~\cite{ma2024groma, jiang2024chatrex}, and (iii) special token decoding~\cite{wang2023visionllm, zhang2024llava}.
Direct coordinate prediction often suffers from error propagation, especially in dense object scenarios.
Detect-and-retrieval methods use bounding boxes as proposals for the MLLM to retrieve candidate regions, posing a new challenge for discriminating among multiple local details.
By comparison, special token decoding offers flexibility by aligning the MLLM with an external decoder.
However, this strategy has mainly focused on segmentation~\cite{zhu2026lens, wang2026x}, leaving its potential for object detection largely under-explored.
Furthermore, such methods force MLLMs to generate special tokens (e.g., <Det>) that lack explicit semantic reference, disrupting their inherent sequential reasoning capabilities.
The absence of intermediate thinking steps also leads to homogeneous outputs among special tokens, hindering adaptation to queries of varying complexity.

To address these limitations, we propose Mixture-of-Thought-Tokens (Motto), a new framework for free-form visual grounding (Figure ~\ref{fig:motivation}), which enables  MLLMs to handle arbitrary queries. Specifically, we propose Spatially-Grounded Thought Tokenization to unify input formulation and region grounding. The image is partitioned into an $N\times N$ grid of regions, where each thought token corresponds to a specific location with a spatial focus. The MLLM generates these thought tokens within the reasoning chain, effectively compressing both visual cues and semantic clues into interpretable latent representations. Then a decoder reconstructs the target regions from these spatially-aligned thought tokens, ensuring that every step is grounded in explicit evidence. 

To effectively bridge the gap between visual perception and semantic reasoning, we propose Context-Adaptive Chain-of-Tokens, which constructs a mixture of thought tokens through a Scan-Focus-Action paradigm. This enables the model to dynamically adapt its grounding mode and integrate diverse evidence based on contextual semantics, achieving precise visual grounding.
In the Perceptual Scanning stage, Perceptual Thought Tokens (P-Tokens) are generated to capture rich visual cues for fine-grained object perception. 
A Switch Adapter then evaluates the aggregated context query to determine whether to perform direct perceptual grounding or activate  reasoning-intensive grounding.
If the model exhibits high confidence, P-Tokens are decoded directly to achieve perceptual grounding.
Otherwise, the pipeline enters the Contextual Re-focusing stage, where textual reflection clues are incorporated to supplement missing semantic information. Following this, the Focus Adapter aggregates candidate visual regions and injects targeted visual evidence to reorient the model toward relevant spatial regions.
Finally, in the Decisive Action stage, the model leverages the semantic and visual contexts to generate Reasoning Thought Tokens (R-Tokens) which are combined with P-Tokens to enable robust reasoning-driven grounding decisions. 

 To measure the performance gap between fine-grained spatial localization and complex semantic reasoning, we construct PR-Bench, a comprehensive referring expression comprehension (REC) benchmark that disentangles referring expressions into these two core dimensions. In summary, our main contributions are as follows: 
\begin{itemize}
    \item We propose Mixture-of-Thought-Tokens (Motto), a unified framework integrating visual perception and semantic reasoning to ground arbitrary queries of varying complexity.  
    \item We design Spatially-Grounded Thought Tokenization to ensure explicit spatial interpretability, and develop Context-Adaptive Chain-of-Tokens, which dynamically switches between different grounding modes to generate thought tokens and evidence in an interleaved chain-of-thought.
    \item We propose PR-Bench, a REC benchmark designed to pinpoint the perception-reasoning gap. Extensive experiments across more than 10 benchmarks show that Motto achieves state-of-the-art performance, establishing a strong baseline for free-form multimodal grounding with MLLMs.
\end{itemize}

\begin{figure}
  \includegraphics[width=1\linewidth]{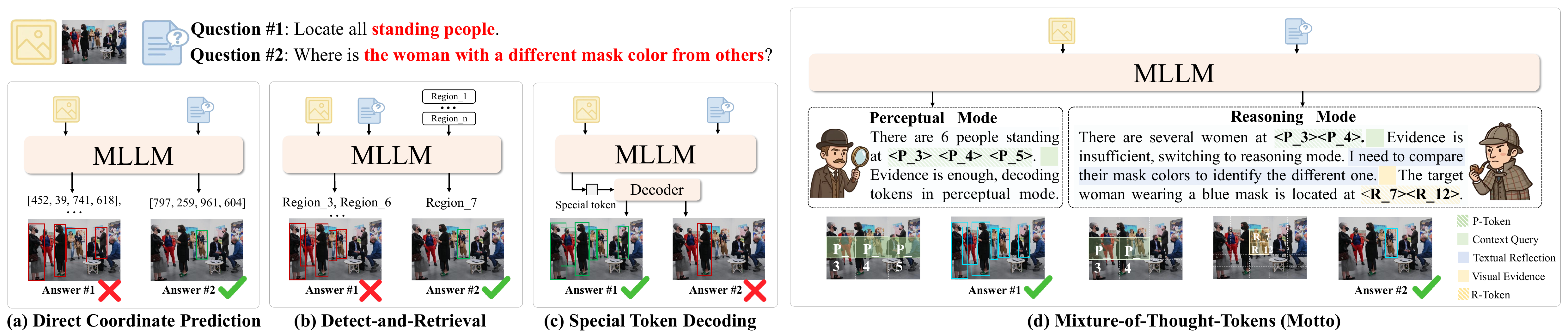}
  \caption{Existing methods struggle to balance precise localization and semantic reasoning, hindering optimal performance. In contrast, Motto generates spatially-aligned thought tokens within an interleaved chain that selects appropriate mode, enabling unified perception and reasoning.}
  \label{fig:motivation}
\end{figure}

\begin{figure}
  \includegraphics[width=\linewidth]{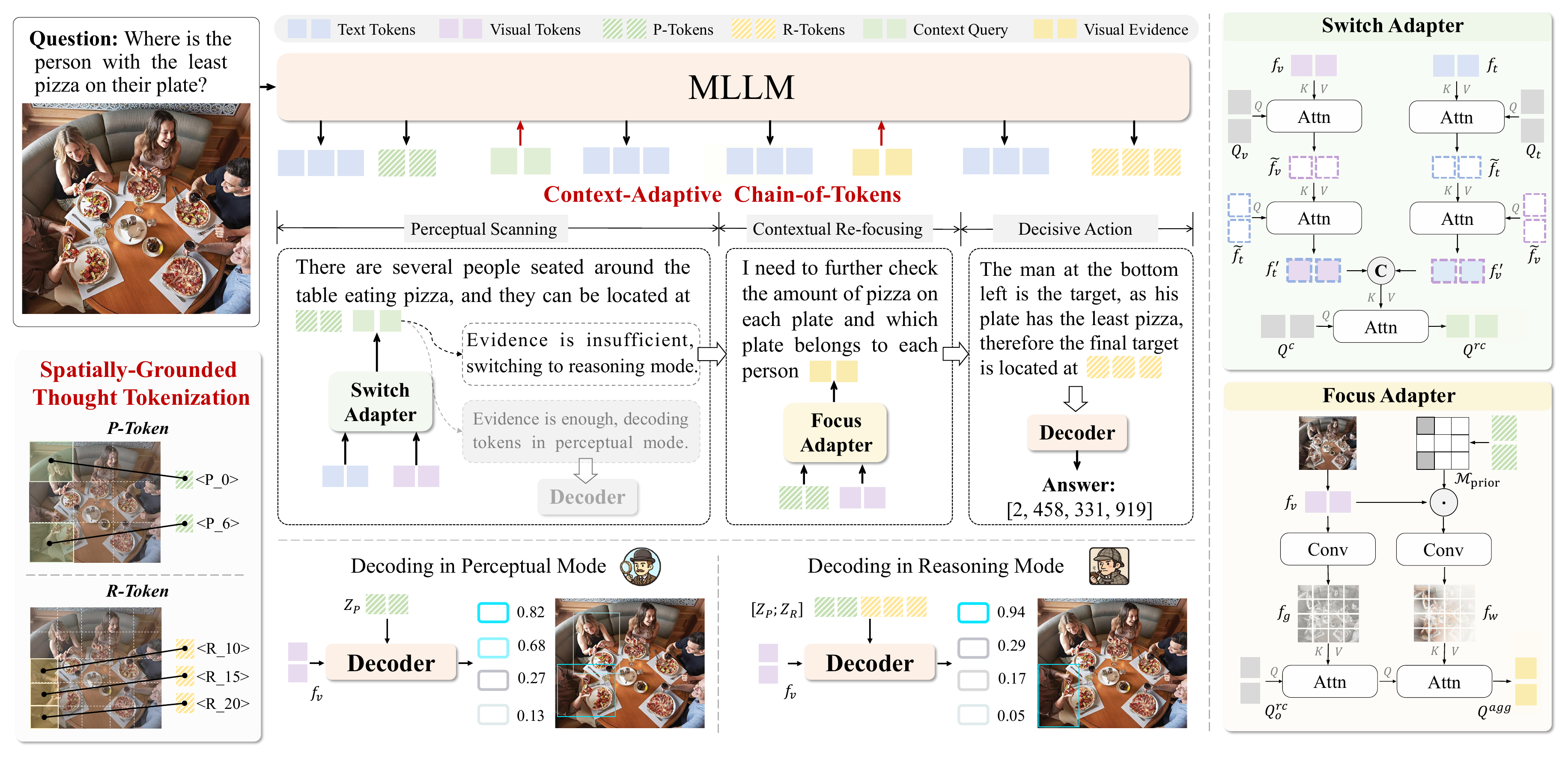}
  \caption{
  Overview of Mixture-of-Thought-Tokens. We propose spatially-grounded thought tokenization to align thought tokens with specific spatial locations. Then Context-Adaptive Chain-of-Tokens is proposed to enable  MLLMs to adaptively switch grounding modes following the Scan-Focus-Action paradigm. 
The model first predicts Perceptual Thought Tokens (P-Tokens) to capture fine-grained visual cues. 
A Switch Adapter then evaluates the context to determine the appropriate grounding mode. For simple queries, P-Tokens are directly decoded by the decoder. However, for complex queries requiring deeper reasoning, textual reflections and refined visual evidence are leveraged as a re-focused context to guide the generation of Reasoning Thought Tokens (R-Tokens), which are then jointly decoded by the decoder along with P-Tokens.
}
  \label{fig:method}
\end{figure}

\section{Related Works}

\subsection{Free-Form Multimodal Grounding}

Free-form multimodal grounding is the task of localizing objects within target images based on diverse queries, encompassing various combinations of modalities and settings. 
Open-vocabulary detection (OVD)~\cite{zareian2021open} enables text-driven detection over open categories, typically evaluated on COCO~\cite{lin2014microsoft} and LVIS~\cite{gupta2019lvis}. 
Referring Expression Comprehension (REC)~\cite{nagaraja2016modeling,xiao2025towards, kang2025visual} focuses on grounding specific regions given a referential sentence, commonly evaluated on the RefCOCO series, including RefCOCO~\cite{yu2016modeling}, RefCOCO+~\cite{yu2016modeling}, and RefCOCOg~\cite{mao2016generation}. 
To address generalized scenarios involving multi-target or no-target cases, more challenging benchmarks have been introduced, such as gRefCOCO~\cite{he2023grec} and HumanRef~\cite{jiang2025referring}.
Recent benchmarks significantly increase query complexity. For example, Ref-L4~\cite{chen2025revisiting} introduces long descriptive expressions and implicit queries requiring visual reasoning, while MC-Bench~\cite{xu2025mc} and MIG-Bench~\cite{li2025migician} extend grounding to multi-image contexts. 
Collectively, these advances shift grounding from simple phrase-to-region matching toward free-form multimodal tasks with deep semantic reasoning~\cite{chen2023advancing, zheng2024resvg,  hu2025groundingsuite, hu2025remerec}, motivating our unified framework that integrates both perceptual and reasoning-based capabilities.

\subsection{MLLMs in Grounding Tasks.}
Existing MLLM-based grounding methods generally follow three strategies. 
The first is direct coordinate prediction. Shikra~\cite{chen2023shikra} predicts coordinates as natural language text, while KOSMOS-2~\cite{peng2023kosmos} represents grounded regions via Markdown-style links containing location tokens. However, this method  suffers from error propagation: even a minor token-level error can corrupt the entire bounding box~\cite{liu2025vlm}. 
The second is detect-and-retrieval. Methods like Groma~\cite{ma2024groma} and ChatRex~\cite{jiang2024chatrex} formulate grounding as region-token retrieval. ROD-MLLM~\cite{yin2025rod} introduces a low-level locator for proposal generation, while VGent~\cite{kang2025vgent} combines these proposals with  hidden states of MLLM for decoder-based selection. Although this strategy avoids direct coordinate generation, it relies heavily on proposal quality and struggles to discriminate fine-grained details across multiple candidate regions. 
The third is special token decoding. 
LLaVA-Grounding~\cite{zhang2024llava} connects MLLM with a grounding model via a special token, while VisionLLM~\cite{wang2023visionllm} and VisionLLM v2~\cite{wu2024visionllm} further adopt task-specific and routing tokens. However, implicit alignment via special tokens often limits reasoning abilities.

\section{Methodology}
We propose a unified framework for free-form multimodal grounding that integrates an MLLM to generate a mixture of thought tokens compressing visual and semantic cues in the latent space, while a decoder maps these representations to  coordinates.

\subsection{Spatially-Grounded Thought Tokenization}

We define \textbf{free-form multimodal grounding} as the task of localizing target regions in response to arbitrary combinations of textual or visual queries, spanning single to multiple candidate images.

We unify all input information into a single sequence $\mathcal{X} = \{\mathcal{I}, q_v, q_t\}$ by leveraging the multimodal encoding capabilities of MLLMs. 
Here, $\mathcal{I}$ denotes one or more target images, while the visual query $q_v$ consists of reference images or spatial cues (e.g., boxes, points) rendered directly onto them. 
The textual query $q_t$ encompasses various forms, ranging from category names to questions involving implicit relations.
Since both $q_v$ and $q_t$ are optional, the model is able to handle arbitrary input combinations. Given this unified input condition, the model autoregressively generates the output sequence $Y = (y_1, \dots, y_L)$ as:
\begin{equation}
    P(Y \mid \mathcal{X}; \theta) = \prod_{i=1}^{L} P(y_i \mid y_{<i}, \mathcal{I}, q_v, q_t; \theta),
\end{equation}
where $\theta$ denotes the learnable parameters.

To unify the output representation,  MLLM first perceives fine-grained image details and performs semantic reasoning by predicting continuous special tokens, enabling perception and reasoning directly within the latent space. 
Second, a decoder predicts precise bounding boxes from these latent representations.
To enhance the interpretability of these tokens, we propose \textit{Spatially-Grounded Thought Tokenization} and introduce spatially-aligned thought tokens. 
Specifically, each token serves as a discrete spatial anchor that indicates the region of interest, explicitly corresponding to a unique location in the image.
We partition the target image $\mathcal{I}$ into an $N \times N$ grid, thereby defining a fine-grained vocabulary of $N^2$ spatial anchors linearly indexed  from $0$ to $N^2-1$:

\begin{equation}
    \mathcal{T}_{\text{thought}} = \{ t_{k} \mid 0 \le k < N^2 \},
\end{equation}
where $t_{k}$ denotes the thought token corresponding to the $k$-th spatial anchor, which maps to a specific location in the grid.
Aligning thought tokens with grid locations enables the model to compress diverse perceptual and semantic cues into a  latent space while ensuring spatial interpretability. 
By leveraging these latent representations, the model retains more comprehensive information than discrete text tokens,  achieving robust image-centric perception and reasoning. 
For the ground-truth bounding boxes $\mathcal{B}_q = \{b_1, \dots, b_M\}$ of each query, we construct a dynamic sequence of thought tokens $\mathcal{S}_{\text{thought}}$ by selecting spatial anchors most relevant to the target regions. 
Specifically, for each grid cell $k$ (where $0 \le k < N^2$) in the $N \times N$ grid, we compute an aggregated overlap score $\alpha_k$. 
This score is defined as $\alpha_k = \sum_{b \in \mathcal{B}_q} \text{IoU}(C_k, b)$, representing the total coverage of all target boxes on that cell, where $C_k$ denotes the  $k$-th grid cell. 
We then rank cells by $\alpha_k$ and select the top-ranked ones as thought tokens, discarding any with zero overlap. 
The number of selected tokens is dynamically determined within the range $[K_{\min}, K_{\max}]$. 
Formally, the sequence is defined as:
\begin{equation}
    \mathcal{S}_{\text{thought}} = \left\{ t_k \in \mathcal{T}_{\text{thought}} \;\middle|\; \text{rank}(\alpha_k) \le K, \; \alpha_k > 0 \right\},
\end{equation}
where $K = \min\big(K_{\max}, \max(K_{\min}, |\{ k \mid \alpha_k > 0 \}|)\big)$. 
This selective mechanism ensures that the MLLM focuses primarily on relevant spatial anchors, enhancing both efficiency and interpretability.

\subsection{Context-Adaptive Chain-of-Tokens}

\noindent\textbf{Perceptual Scanning.}
Given an input query, the MLLM first identifies the target image from the candidate set if multiple images are provided. It then generates a sequence of perceptual thought tokens on an $N^p \times N^p$ grid, with length adaptively varying within $[K_{\min}^{p}, K_{\max}^{p}]$, to capture fine-grained perceptual details within the latent representations.
The generated sequence, denoted as $\mathcal{S}^p$, is then processed by a Switch Adapter, which summarizes query-relevant information to produce an in-context indicator for mode selection.
Specifically, we compress the image features $f_v \in \mathbb{R}^{N_v \times d}$ and text embeddings $f_t \in \mathbb{R}^{N_t \times d}$ using learnable queries $Q_v \in \mathbb{R}^{L_v \times d}$ and $Q_t \in \mathbb{R}^{L_t \times d}$, where $L_v$ and $L_t$ denote the number of queries and $d$ is the shared  dimension. This is followed by a cross-modal fusion:
\begin{equation}
    \begin{aligned}
        \tilde{f}_v &= \mathrm{Attn}(Q_v, f_v, f_v), \quad \tilde{f}_t = \mathrm{Attn}(Q_t, f_t, f_t), \\
        f'_t &= \mathrm{Attn}(\tilde{f}_t, \tilde{f}_v, \tilde{f}_v), \quad f'_v = \mathrm{Attn}(\tilde{f}_v, \tilde{f}_t, \tilde{f}_t).
    \end{aligned}
\end{equation}
Subsequently, the fused visual and textual features ($f'_v$ and $f'_t$) are combined to update the global learnable context queries $Q^c \in \mathbb{R}^{L \times d}$:
\begin{equation}
    Q^{rc} = \mathrm{Attn}(Q^{c}, [f'_t; f'_v], [f'_t; f'_v]),
\end{equation}
where  $[\cdot;\cdot]$ indicates feature-wise concatenation.
This refined context query $Q^{rc}$ is appended after $\mathcal{S}^p$. Conditioned on switch context $\mathcal{X}^{\text{sw}}$, MLLM then  generates a decision sequence $Y$ of length $L_{\text{sw}}$:
\begin{equation}
    P(Y \mid \mathcal{X}^{\text{sw}}; \theta)
    =
    \prod_{i=1}^{L_{\text{sw}}}
    P\!\left(y_i \mid y_{<i}, \mathcal{X}, T^{\text{id}}, \mathcal{S}^p, Q^{rc}; \theta\right).
\end{equation}
where $T^{\text{id}}$ denotes the text that indicates the target image. 
MLLM determines whether visual cues are sufficient. When visual cues are sufficient,  MLLM generates a trigger sequence (e.g., `decode tokens in perceptual mode') to terminate reasoning and execute decoding. Otherwise, the process advances to reasoning mode.

\noindent\textbf{Contextual Re-focusing.}
In the reasoning mode, MLLM first executes Contextual Re-focusing by combining textual reflections and targeted visual evidence. 
First, MLLM generates explicit textual reflection chains to identify limitations in the latent representations captured by P-Tokens, avoiding the direct projection of visual cues into the discrete text space. 
This strategy mitigates the loss of subtle perceptual details inherent in converting continuous representations into discrete symbols. 
Instead, these reflections explicitly characterize grounding challenges, such as broad target distributions, small object scales, or semantic ambiguities,  ensuring reasoning coherence while preserving the continuity of the latent space.
Subsequently, a Focus Adapter is employed to summarize candidate visual regions identified by the P-Tokens. 
We first construct a visual patch map $\mathcal{M}_{\text{prior}}$ based on the grid locations indicated by P-Tokens. Specifically, each patch from the visual features $f_v$ is assigned a value of 1 if it  overlaps with any selected grid region, and 0 otherwise. This binary map effectively highlights  patches relevant to the target distribution. The visual features $f_v$ are then processed to capture both global and target-focused contexts.
\begin{equation}
    f_{g} = \mathrm{Conv}_{g}(f_{v}), \quad f_{w} = \mathrm{Conv}_{w}(f_{v} \odot \mathcal{M}_{\text{prior}}),
\end{equation}
where $f_g$ and $f_w$ denote  global and target-focused features, and $\odot$ denotes the element-wise product.
To integrate the perceptual context from the generated P-Tokens, we utilize the output representation of the context query $Q^{rc}$ from the MLLM, denoted as $Q^{rc}_o$, as the initialized query and perform hierarchical cross-attention:
\begin{equation}
    Q^{\text{agg}} = \mathrm{Attn}\big(\mathrm{Attn}(Q^{rc}_o, f_{g}, f_{g}), f_{w}, f_{w}\big),
\end{equation}
which  transforms the queries into compact zoom-in visual tokens by first incorporating global features and then refining them with target-specific ones. 
These resulting queries $Q^{\text{agg}}$ serve as re-focused visual evidence and are appended after the textual reflections to jointly guide the generation of the reasoning thought tokens.

\noindent\textbf{Decisive Action.}
Building on the textual reflections and visual evidence aggregated in the previous stages, the MLLM enters the Decisive Action stage to generate a dynamic sequence of Reasoning Thought Tokens (R-Tokens). The probability is formulated as:
\begin{equation}
    P(Y \mid \mathcal{X}^{\text{r}}) = \prod_{i=1}^{L_{\text{r}}} P(y_i \mid y_{<i}, \mathcal{X}^{\text{sw}}, T^{\text{ref}}, Q^{\text{agg}}; \theta),
\end{equation}
where $T^{\text{ref}}$ and $Q^{\text{agg}}$ represent the textual reflections and visual evidence. The generated sequence, denoted as $\mathcal{S}^r$, operates on a grid of size $N^r \times N^r$, with its length $L_{\text{r}}$ adaptively varying within $[K_{\min}^{r}, K_{\max}^{r}]$. 
We set $N^r > N^p$ to achieve a finer grid partition, ensuring that each token corresponds to a smaller  grid for more precise localization.
Subsequently, a DETR-like~\cite{zhu2020deformable} decoder is introduced to convert these spatially-aligned thought tokens into explicit regions. 
Comprising three Transformer decoder layers and two prediction heads for bounding box regression and confidence scoring, the decoder initializes object queries by projecting latent representations, conditioned on the grounding mode:
\begin{equation}
    Q^{\mathrm{object}} =
    \begin{cases}
        \mathrm{MLP}(Z_P), & \text{if mode} = \text{perceptual}, \\
        \mathrm{MLP}([Z_P; Z_R]), & \text{if mode} = \text{reasoning},
    \end{cases}
\end{equation}
where $Z_P$ and $Z_R$ denote the latent representations of P-Tokens and R-Tokens, respectively, $[\cdot;\cdot]$ indicates concatenation along the token dimension, and $\mathrm{MLP}(\cdot)$ projects the representations into the decoder's query space. 
Finally, the modeled object queries $Q^{\mathrm{object}}$ interact with the target image features through the Transformer layers to produce the final bounding boxes and confidence scores.
\subsection{Training Strategy}

\noindent\textbf{Data Construction Strategy.}
To facilitate adaptive mode selection based on query difficulty, we propose a bi-mode data construction framework that operates on two  data pools:
(i) Open-Vocabulary Detection Data: We implement a rule-based hard mining strategy. Instances from tail classes are directly identified as hard samples. For head categories, we analyze target density and bounding box aspect ratios. Queries exhibiting high target density or tiny   bounding box areas are classified as hard samples, while others are treated as easy samples. 
(ii) Referring Grounding Data: We employ Qwen3-VL-235B-A22B~\cite{bai2025qwen3} to evaluate query difficulty via 8 repeated generations per query. Queries successfully resolved in over 50\% of attempts are categorized as easy samples. 
For the remaining candidates, we adopt the same MLLM to perform a secondary refinement: those that correctly identify the target despite slight localization deviations are reclassified as easy samples, whereas others exhibiting core reasoning failures or missing critical evidence remain classified as hard samples.
Based on this classification, we construct ground-truth sequences. For easy samples, we utilize P-Tokens to directly encode target regions. For hard samples, we employ the MLLM to explicitly reflect on error causes and summarize them into a reflective reasoning chain.  These hard samples are then formed by combining P-Tokens with the R-Tokens. Detailed distribution statistics are provided in the Supplementary Material.

\noindent\textbf{Training Stages.} 
Our training strategy consists of two stages to balance perceptual and reasoning capabilities. 
First, we pre-train the model  on easy open-vocabulary detection samples, focusing on learning perceptual scanning and generating P-Tokens to enhance generalization to diverse object categories. 
Second, we fine-tune the model on the full dataset to cover the complete Scan-Focus-Action paradigm. This enables the model to distinguish between grounding modes and leverage both P-Tokens and R-Tokens for reasoning-based decisions, while retaining fine-grained detection.



\noindent\textbf{Training Objectives.} 
We optimize the model using two main objectives: a language modeling loss $\mathcal{L}_{\text{txt}}$ and a grounding loss $\mathcal{L}_{\text{det}}$. For the autoregressive output, the language loss employs standard supervised fine-tuning with per-token cross-entropy:
\begin{equation}
    \mathcal{L}_{\text{txt}} = \frac{1}{T}\sum_{t} -\log p(\hat{y}_t \mid \mathcal{X}, y_{<t}),
\end{equation}
where $\hat{y}_t$ denotes the ground-truth token at step $t$. The grounding loss $\mathcal{L}_{\text{det}}$ comprises box regression and classification components, where predicted boxes are matched to ground-truth boxes via the Hungarian algorithm~\cite{kuhn1955hungarian}. Following prior works~\cite{carion2020end,meng2021conditional,liu2022dab}, we utilize L1 loss $\mathcal{L}_{\text{L1}}$ and GIoU loss $\mathcal{L}_{\text{giou}}$~\cite{rezatofighi2019generalized} for box regression. For classification, we adopt the focal loss strategy from Grounding DINO~\cite{liu2024grounding} with a key modification such that we retain all positive predictions while applying focal-loss supervision exclusively to the top-$K$ highest-scoring hard negatives as formulated:
\begin{equation}
    \mathcal{L}_{\mathrm{cls}} = \frac{1}{N_{\mathrm{sel}}} \sum_i \left[ \sum_{j\in \mathcal{P}_i} \mathrm{FL}(\hat{s}_{i,j}, 1) + \sum_{j\in \mathcal{N}_{i}^{\mathrm{hard}}} \mathrm{FL}(\hat{s}_{i,j}, 0) \right],
\end{equation}
where \(i\) and \(j\) denote the sample index and prediction index, \(\mathrm{FL}(\cdot)\) denotes the focal loss, and \(\hat{s}_{i,j}\) denotes the predicted classification score of the \(j\)-th prediction in the \(i\)-th sample. \(\mathcal{P}_i\) and \(\mathcal{N}_i^{\mathrm{hard}}\) denote the sets of matched positive predictions and top-$K$ highest-scoring hard negatives for the \(i\)-th sample, respectively, and \(N_{\mathrm{sel}}=\sum_i (|\mathcal{P}_i|+|\mathcal{N}_i^{\mathrm{hard}}|)\) represents the total number of selected predictions across all training samples.
Finally, the overall training objectives combine these components as follows:
\begin{equation}
    \mathcal{L} = \lambda_{\text{txt}}\mathcal{L}_{\text{txt}} + \lambda_{\text{L1}} \mathcal{L}_{\text{L1}} + \lambda_{\text{giou}} \mathcal{L}_{\text{giou}} + \lambda_{\text{cls}} \mathcal{L}_{\text{cls}},
\end{equation}
where $\lambda_{\text{txt}}$, $\lambda_{\text{L1}}$, $\lambda_{\text{giou}}$, and $\lambda_{\text{cls}}$ are balancing weights for the respective loss terms of language modeling and grounding.

\subsection{Statistics of PR-Bench}
We construct \textbf{PR-Bench}, a comprehensive REC benchmark evaluating two core capabilities: \textit{Visual-cue Perception} and \textit{Compositional Reasoning}. 
\textit{Visual-cue Perception} comprises three subcategories: (1) Attribute, capturing intrinsic visual properties; (2) Position, defining spatial relationships among objects; and (3) Interaction, describing relative relationships within categories. 
\textit{Compositional Reasoning} addresses more complex tasks: (4) Relation, involving multi-object compositional referring; (5) Commonsense, requiring contextual identification without explicit naming; and (6) Rejection, handling queries for absent objects.
We construct PR-Bench using images from FineHARD~\cite{xie2025fg}.  An MLLM-based annotation pipeline is proposed to generate initial query–box pairs, which are then refined by ten human annotators. 
PR-Bench contains 6,000 evenly distributed pairs across six subcategories, utilizing high-resolution images of over 1,000 object types. It maintains an average target area ratio of 9\%, with a specific emphasis on marginally visible targets.

\section{Experiment}
\subsection{Experimental Setup}

\noindent\textbf{Datasets and Tasks.}
We construct our training data pool, including Object365~\cite{shao2019objects365} as open-vocabulary detection data, and a diverse set of referring grounding data including MGrounding-630k~\cite{li2025migician}, RefCOCO/+/g~\cite{yu2016modeling,mao2016generation}, VisualGenome~\cite{krishna2017visual}, gRefCOCO~\cite{he2023grec}, HumanRef~\cite{jiang2025referring}, ReasonSeg~\cite{lai2024lisa}, OmniRef~\cite{zheng2025omni} and Rexverse2M~\cite{jiang2024chatrex}.
We evaluate Motto across three core tasks.
For open-vocabulary detection, we follow T-Rex2~\cite{jiang2024t} to report results on COCO~\cite{lin2014microsoft} across three prompt settings: Text (category name), Visual-G (reference image), and Visual-I (interactive box).
For REC, we assess performance on RefCOCO~\cite{yu2016modeling}, RefCOCO+~\cite{yu2016modeling}, RefCOCOg~\cite{mao2016generation}, Ref-L4~\cite{chen2025revisiting}, gRefCOCO~\cite{he2023grec},  HumanRef~\cite{jiang2025referring} and ReasonSeg~\cite{lai2024lisa}, with inputs ranging from short phrases and long captions to reasoning questions, and targets from single to multiple objects. 
Note that ReasonSeg \cite{lai2024lisa} is utilized by extracting bounding boxes from  original  masks. 
Then we adopt MIG-Bench~\cite{li2025migician} for  multi-image grounding.
We further evaluate the model's comprehensive perception and reasoning capabilities on the proposed PR-Bench.

\noindent\textbf{Evaluation Metrics.}
For REC, following standard protocols~\cite{kazemzadeh2014referitgame,yu2016modeling,mao2016generation}, we employ Acc@0.5 for single-object grounding on RefCOCO~\cite{yu2016modeling}, RefCOCO+~\cite{yu2016modeling}, RefCOCOg~\cite{mao2016generation}, and Ref-L4~\cite{chen2025revisiting}. 
For multi-object grounding, we report Pr@\mbox{$(F1=1,\ \mathrm{IoU}\geq 0.5)$} (Pr)  and No-Target Accuracy (N-Acc) on gRefCOCO~\cite{he2023grec} and ReasonSeg~\cite{lai2024lisa}, as well as Density-F1 (DF1) and Rejection Score (Rej) on HumanRef~\cite{jiang2025referring}. 
In multi-image grounding, we utilize Acc@0.5 on MIG-Bench~\cite{li2025migician}. 
For object detection on COCO~\cite{lin2014microsoft}, we report mAP and F1, where F1 denotes F1@mIoU.
For PR-Bench, we adopt Mean Accuracy (mAcc) and N-Acc. The mAcc is calculated as the average accuracy across the five non-rejection subcategories, with scores averaged over IoU thresholds from 0.5 to 0.9 (step 0.05). 
Additionally, we report two specialized metrics: $\text{mAcc}_{API}$, averaged over the three subcategories in Visual-Cue Perception, and $\text{mAcc}_{RC}$, averaged over the two subcategories in Compositional Reasoning excluding Rejection.

\noindent\textbf{Implementation Details.} 
Motto is built upon the Qwen3-VL-2B~\cite{bai2025qwen3}. 
In the first stage, we train all parameters for 2 epochs with a batch size of 512. The learning rates are set to $1 \times 10^{-4}$ for decoder and $1 \times 10^{-5}$ for MLLM. 
In the second stage, we train for 2 epochs with a batch size of 128, adopting learning rates of $2 \times 10^{-5}$ for decoder and $1 \times 10^{-5}$ for MLLM. 
The grid sizes for P-tokens and R-tokens are set to $3 \times 3$ and $5 \times 5$, introducing a total of 34 extra tokens into the textual vocabulary, with maximum token lengths of 4 for P-tokens and 6 for R-tokens. 
For the loss function, the weighting coefficients are configured as $\lambda_{\text{txt}}=1.0$, $\lambda_{\text{L1}}=2.0$, $\lambda_{\text{giou}}=5.0$, and $\lambda_{\text{cls}}=4.0$. We set $K=20$ for top-$K$ hard negative selection in $\mathcal{L}_{\text{cls}}$. 
All experiments are conducted on 32 NVIDIA A800-80G GPUs.

\newcommand{\best}[1]{\textbf{#1}}
\newcommand{\second}[1]{\underline{#1}}

\begin{table*}
  \centering
  \caption{Comparisons on general referring expression comprehension benchmarks. Best results are shown in \textbf{bold}, and the second-best are \underline{underlined}. “--” indicates unavailable or unreported results.}
  \label{tab:exp_rec}

  \resizebox{\textwidth}{!}{%
    \begin{tabular}{l c c c c c c c c c c c c}
\toprule
\multirow{2}{*}{\textbf{Model}} & \multirow{2}{*}{\textbf{Params.}} & \multirow{2}{*}{\textbf{RefCOCO}}
& \multirow{2}{*}{\textbf{RefCOCO+}}
& \multirow{2}{*}{\textbf{RefCOCOg}}
& \multirow{2}{*}{\textbf{Ref-L4}}
& \multicolumn{2}{c}{\textbf{gRefCOCO}}
& \multicolumn{2}{c}{\textbf{HumanRef}}

& \multirow{2}{*}{\textbf{ReasonSeg}}
& \multicolumn{2}{c}{\textbf{PR-Bench}}\\
\cmidrule(lr){7-8}\cmidrule(lr){9-10}\cmidrule(lr){12-13}
& 
& 
& 
& 
& 
& Pr & N-Acc
& DF1 & Rej
 &  & mAcc & N-Acc\\
\midrule
  \multicolumn{13}{l}{\textbf{\textit{General MLLMs}}} \\
Qwen3-VL~\cite{bai2025qwen3} &2B &87.4  &78.1  &84.9 &85.4  &46.5  &25.1  &58.9  &--  &49.8   &52.8 &--  \\
Youtu-VL~\cite{wei2026youtu} &4B &\underline{93.2} &\underline{89.8} &\underline{92.6}  &83.3 &25.7  &--  &55.3  &-- &51.3   &53.2 &-- \\

InternVL3.5~\cite{wang2025internvl3} &8B &91.9   &87.6   &89.5  &80.9  &31.2  &--  &40.0  &33.1   &38.3  &41.5 &--  \\

Qwen3-VL~\cite{bai2025qwen3} &8B &90.6  &84.8  &88.2  &88.5  &55.4  &55.0  &64.9  &\underline{47.9} &56.4   &59.4 &\underline{16.2}  \\

GLM-4.1V-Thinking~\cite{hong2025glm} & 9B &88.9 &82.9 &86.3 &84.8 &49.3 &29.0 &58.6 &34.5 &57.6 &57.9 &-- \\
Qwen3.5~\cite{qwen3.5} &9B &87.2  &88.8  &88.2  &89.0 &56.3  &--  &72.0  &10.8  &56.6   &63.7 &11.9 \\
Qwen3.5~\cite{qwen3.5} &27B &92.5 &89.6 &90.6 &\underline{90.2} &56.7 &-- &71.5 &13.4 &65.7 &\underline{65.3} &-- \\
InternVL3.5~\cite{wang2025internvl3} &38B &90.4 &87.4 &89.8 &85.5 &36.0 &-- &51.1 &-- &51.7 &48.5 &12.7 \\

\midrule

\multicolumn{13}{l}{\textbf{\textit{Grounding-Specialized MLLMs}}} \\
VLM-R1~\cite{shen2025vlm} &3B &89.2  &83.5  &86.2 &80.9  &45.7  &30.6  &46.4  &27.1  &26.5  &54.4 &--  \\
VLM-FO1~\cite{liu2025vlm} &3B &90.8   &86.3   &88.6 &84.4    &\underline{58.7}  &52.5   &\underline{82.6} &46.4  &46.7   &53.9 &11.4  \\
Rex-Omni~\cite{jiang2025detect}{\scriptsize[CVPR'26]} &3B &86.3  &78.6  &86.7  &78.0 &38.0  &33.6  &74.9  &--  &40.6   &53.7 &--  \\

Chatrex~\cite{jiang2024chatrex} &7B &90.7   &87.0   &89.9   &80.0 &29.9  &--  &55.6 &--  &23.0   &49.5 &--  \\
ROD-MLLM~\cite{yin2025rod}{\scriptsize[CVPR'25]} &7B &89.8 &84.1 &86.7 &-- &53.5 &\underline{57.8   } &--&--&--&--&--\\

Migician~\cite{li2025migician}{\scriptsize[ACL'25]} &7B &90.8   &85.7   &87.9 &73.2  &29.4  &14.4  &45.9  &26.0   &47.3   &52.3 &--  \\
DeepEyes~\cite{zheng2025deepeyes}{\scriptsize[ICLR'26]} &7B &89.8  &83.6  &86.7 & 85.2 &50.3 &38.4 &50.3 &22.8 &\underline{68.6}  &45.7 &--\\
VGent~\cite{kang2025vgent}{\scriptsize[CVPR'26]} &7B &92.3  &87.9  &90.3  &--&--&--&--&--&--&--&--  \\

\rowcolor{gray!10}  \textbf{Motto}& \textbf{2B}
      & \textbf{94.0} & \textbf{90.8} & \textbf{93.1} & \textbf{92.5}& \textbf{69.6} & \textbf{58.0} & \textbf{83.0} & \textbf{53.4}  &\textbf{71.2} & \textbf{71.7} & \textbf{46.9} \\
    \bottomrule
  \end{tabular}%
  }
\end{table*}

\begin{table}[t]
    \centering

    \begin{minipage}[t]{0.44\textwidth}
        \vspace{0pt}
        \caption{Comparison on COCO under different prompt settings (Text, Visual-G, and Visual-I). “--” indicates unavailable results, and  $^{*}$ denotes non-zero-shot results. Best and second-best results are highlighted in \textbf{bold} and \underline{underlined}.} 
        \label{tab:prompt_type_results_reformatted}

        \scriptsize
        \setlength{\tabcolsep}{4pt}
        \renewcommand{\arraystretch}{1.31}
        \resizebox{\linewidth}{!}{
        \begin{tabular}{lcccccc}
            \toprule
            \multirow{2}{*}{\textbf{Model}} & \multicolumn{2}{c}{\textbf{Text}} & \multicolumn{2}{c}{\textbf{Visual-G}} & \multicolumn{2}{c}{\textbf{Visual-I}} \\
            \cmidrule(lr){2-3} \cmidrule(lr){4-5} \cmidrule(lr){6-7}
             & mAP & F1 & mAP & F1 & mAP & F1 \\
            \midrule
            T-Rex-2 & \underline{52.2} & --   & \underline{46.5} & --   & \underline{58.5} & --   \\
            VLM-FO1-3B  & 44.4$^{*}$ & --   & 14.1$^{*}$ & 24.9$^{*}$ & 11.0$^{*}$   & 20.3$^{*}$ \\
            Rex-Omni-3B & --   & \underline{52.9} & 17.4 & 29.9& -- & \underline{61.3}  \\
            Youtu-VL-4B &47.1 &-- &18.4 &26.8  &21.0 &27.4 \\
            Qwen3.5-9B    & 37.7 & 48.7 & 20.8 & 36.9 & 32.6 & 45.1 \\
            Qwen3.5-27B     & 38.8 & 50.6 & 22.2 & \underline{37.4} & 33.3 & 45.6 \\
            \rowcolor{gray!10} \textbf{Motto}   & \textbf{56.4} & \textbf{62.1}  & \textbf{51.7} & \textbf{57.4} & \textbf{58.9} & \textbf{65.7} \\
            \bottomrule
        \end{tabular}}
    \end{minipage}
    \hfill
    \begin{minipage}[t]{0.54\textwidth}
        \vspace{0pt}
        \caption{Comparison on MIG-Bench~\cite{li2025migician}. 
         \textbf{SG} (Spontaneous Grounding) averages results across Static, Robust, and Common settings. 
        \textbf{V} (Visual Reference) aggregates Object Tracking (OT), Multi-View (MV), Region, and Refer tasks. 
        \textbf{T} (Textual) corresponds to Group Grounding (GG). 
         \textbf{V+T} averages Reasoning and Correspondence (Co-Re) tasks. 
        Best and second-best results are highlighted in \textbf{bold} and \underline{underlined}.}
        \label{tab:regroup_results}

        \scriptsize
        \setlength{\tabcolsep}{4pt}
        \renewcommand{\arraystretch}{1.00}
        \resizebox{\linewidth}{!}{
        \begin{tabular}{lcccccc}
            \toprule
            \textbf{Model} & \textbf{Params.} & \textbf{SG} & \textbf{V} & \textbf{T} & \textbf{V+T} & \textbf{AVE} \\
            \midrule
            Youtu-VL &4B & 33.3 & 37.3 & 37.9 & 21.1 & 32.9 \\
            Migician   &7B  &\underline{65.4} &\underline{70.5} &\underline{66.5} &46.8 &\underline{63.8} \\
            Qwen3-VL &8B & 34.6 & 25.6 & 53.9 & 29.2 & 31.8 \\
            Qwen3.5&9B & 40.1 & 29.9 & 44.9 & 16.2 & 31.7 \\
            GLM-4.1V-Thinking& 9B &50.1 &39.5 &63.5 &\underline{47.5} &46.7\\
            Qwen3.5 &27B & 48.7 & 37.8 & 66.1 & 31.9 & 42.7 \\
              \rowcolor{gray!10}\textbf{Motto}     &\textbf{2B}    &\textbf{71.0} & \textbf{81.2} & \textbf{85.4} & \textbf{61.0} & \textbf{74.5} \\
            \bottomrule
        \end{tabular}}
    \end{minipage}
\end{table}

\subsection{Main Results}



\noindent \textbf{Object Detection Under Text and Visual Prompt.} 
We evaluate Motto's zero-shot capabilities on the COCO benchmark. By predicting spatially-aligned thought tokens, Motto explicitly localizes regions of interest while compressing rich perceptual and semantic information into a compact latent space. 
In the standard text prompt setting, Motto achieves competitive performance against existing MLLM methods (e.g., Rex-Omni), despite utilizing only half the training data. 
For visual prompts, we consider two scenarios: image-based queries (Visual-G) and region-specific queries (Visual-I). Motto excels in both, surpassing the second-best model by mAP margins of 5.2\% and 0.4\%, respectively. 
Notably, while general MLLMs already have the capability to handle visual prompts, Motto significantly strengthens this for precise grounding.
These improvements demonstrate strong generalization and the potential of Motto to benefit from larger-scale datasets.

\noindent \textbf{General Referring Expression Comprehension.} 
Table~\ref{tab:exp_rec} compares Motto with state-of-the-art models across diverse benchmarks, avoiding the bias of relying only on RefCOCO/+/g. 
Motto outperforms existing methods on both RefCOCO/+/g (short phrase) and Ref-L4 (long caption), effectively handling queries ranging from short phrases to complex sentences. 
On multi-object benchmarks such as gRefCOCO and HumanRef, Motto accurately localizes multiple targets and confidently rejects queries when no object exists, mitigating the hallucinations common in methods trained only on single positive object. 
Notably, retrieval-based methods such as ChatRex~\cite{jiang2024chatrex} struggle on ReasonSeg due to limited reasoning capabilities, as this task requires deep semantic understanding of the target. In contrast, Motto achieves competitive performance by aggregating semantic and visual contexts within  context-adaptive chain-of-tokens.
Finally, while models saturate on standard datasets like RefCOCO, their performance drops significantly on PR-Bench, highlighting the necessity of combining advanced perception with reasoning.
More detailed comparisons across six subcategories in PR-Bench are illustrated in the supplementary material.

\noindent \textbf{Multi-image Grounding.} 
As shown in Table~\ref{tab:regroup_results}, Motto achieves state-of-the-art performance across all tasks on MIG-Bench, yielding an average improvement of 10.7\% over the second-best model despite having significantly fewer parameters. 
Free-form multi-image grounding remains a major challenge: traditional detection models lack support for such tasks, while existing MLLMs, despite their free-form grounding capabilities, struggle when extended to multi-image scenarios.
Even advanced models like Qwen3.5 underperform, particularly in Visual Reference (V) and Reasoning and Correspondence (V+T). 
In contrast, Motto enables robust multi-image grounding by unifying input formulations and output representations.
This design empowers the model to accurately identify the target image within a candidate set and localize specific regions, by leveraging the cooperation between P-Tokens and R-Tokens.

\subsection{Ablation Studies}

We conduct detailed ablation studies on a 3M sample subset, adopting the same hyperparameters as the full-dataset training.

\noindent\textbf{Effects of Mixture-of-Thought-Tokens.}
As shown in Table~\ref{tab:incremental_ablation_visual_thoughts}, the first two rows serve as baselines using naive special tokens (e.g., \texttt{<Det>}) lacking explicit spatial anchors. In contrast, our P-Tokens and P+R-Tokens settings employ spatially-aligned thought tokens with identical sequence lengths. 
Adopting spatially-aligned thought tokens significantly boosts performance, particularly in the long-token setting. 
Without explicit spatial alignment, increasing the number of special tokens often yields homogeneous outputs, leading to a substantial performance drop. 
Conversely, our P-Tokens and R-Tokens leverage a coarse-to-fine grid partitioning predicted within the reasoning chain. This mechanism sequentially compresses visual cues and semantic clues into interpretable latent representations, thereby achieving superior accuracy. 
Moreover, the Switch Adapter yields an average gain of 1.0\%. This indicates that adaptive selection not only enhances inter-mode cooperation, but also ensures specialized focus: P-Tokens capture comprehensive context, while R-Tokens handle complex reasoning.
Finally, incorporating textual reflection and visual evidence strengthens reasoning capabilities while preserving fine-grained perception, validating the effectiveness of our Mixture-of-Thought-Tokens.

\begin{table}[t]
    \centering

    \begin{minipage}[t]{0.48\textwidth}
        \vspace{0pt}
        \caption{
        Ablation studies on incremental components across COCO, PR-Bench, and MIG-Bench.
        }
        \label{tab:incremental_ablation_visual_thoughts}

        \scriptsize
        \setlength{\tabcolsep}{4pt}
        \renewcommand{\arraystretch}{1.12}
        \resizebox{\linewidth}{!}{
        \begin{tabular}{lcccccc}
        \toprule
        \textbf{Setting} & \textbf{COCO} & \multicolumn{3}{c}{\textbf{PR-Bench}} & \textbf{MIG-Bench} & \textbf{AVG} \\
        & & mAcc$_{API}$ & mAcc$_{RC}$ &N-Acc & & \\
        \midrule
        \multicolumn{7}{l}{\textbf{\textit{Baselines}}} \\
        <Det> (short length) & 44.6 & 67.3 & 54.5 & 45.3 & 58.7 & 54.1 \\
        <Det> (long length)  & 43.2 & 66.8 & 54.2 & 44.7 & 58.5 & 53.5 \\
        \midrule
        \multicolumn{7}{l}{\textbf{\textit{Ours}}}  \\
        P-Tokens             & 46.2 & 69.1 & 55.9 & 46.7 & 62.6 & 56.1 \\
        + R-Tokens           & 45.9 & 70.3 & 59.4 & 45.8 & 65.3 & 57.3 \\
        + Switch Adapter     & 46.4 & 71.6 & 60.9 & 46.5 & 66.2 & 58.3 \\
        + Textual Reflection &46.7 & 72.4 & 61.7 & 46.4 & 66.6 & 58.8 \\
        + Focus Adapter      & 46.6 & 73.2 & 63.2 & 46.3 & 67.5 & 59.4 \\
        \bottomrule
        \end{tabular}}
    \end{minipage}
    \hfill
    \begin{minipage}[t]{0.5\textwidth}
        \vspace{0pt}
        \caption{Ablation studies on different grid sizes in Spatially-Grounded Thought Tokenization.}
        \label{tab:ablation_coordinate_tokens}

        \scriptsize
        \setlength{\tabcolsep}{4pt}
        \renewcommand{\arraystretch}{1.20}
        \resizebox{\linewidth}{!}{
        \begin{tabular}{cc c ccc c c}
        \toprule
        \multicolumn{2}{c}{\textbf{Grid Size}} & \multirow{2}{*}{\textbf{COCO}} & \multicolumn{3}{c}{\textbf{PR-Bench}} & \multirow{2}{*}{\textbf{MIG-Bench}} & \multirow{2}{*}{\textbf{AVG}} \\
        \cmidrule(lr){1-2} \cmidrule(lr){4-6}
        \textbf{P-Token} & \textbf{R-Token} & & mAcc$_{API}$ & mAcc$_{RC}$ & N-Acc & & \\
        \midrule
        $2\times2$ & --          & 45.1 & 68.4 & 55.2 & 46.5 & 61.2 & 55.3 \\
        $3\times3$ & --          & 46.2 & 69.1 & 55.9 & 46.7 & 62.6 & 56.1 \\
        $4\times4$ & --          & 44.8 & 67.9 & 54.4 & 45.4 & 60.8 & 54.7 \\
        $3\times3$ & $3\times3$ & 46.0 & 70.1 & 58.6 & 46.4 & 64.7 & 57.2 \\
        $3\times3$ & $4\times4$ & 46.2 & 72.5 & 61.0 & 46.1 & 66.1 & 58.4 \\
        $3\times3$ & $5\times5$ & 46.6 & 73.2 & 63.2 & 46.3 & 67.5 & 59.4 \\
        $3\times3$ & $6\times6$ & 45.2 & 71.8 & 60.3 & 46.2 & 65.6 & 57.8 \\
        \bottomrule
        \end{tabular}}
    \end{minipage}
\end{table}

\noindent\textbf{Grid Size in Spatially-Grounded Thought Tokenization.}
We explore the effects of varying grid sizes for thought tokens, evaluating both the P-Token-only setting (perceptual mode) and full method combining P- and R-Tokens. 
As shown in Table~\ref{tab:ablation_coordinate_tokens}, the $3\times3$ grid yields optimal results for P-Tokens, especially on COCO. 
This suggests that overly coarse partitions lack sufficient detail, while excessively fine grids fragment visual evidence and restrict the MLLM's receptive field, hindering global context perception. 
Conversely, increasing the R-Token grid size from $3\times3$ to $5\times5$ consistently boosts reasoning metrics, with the hybrid $3\times3$ (P) / $5\times5$ (R) configuration achieving peak performance. 
This indicates that while a coarser grid is sufficient for broad visual scanning, a finer partition in the reasoning stage enables precise localization of subtle details based on the aggregated context.



\begin{figure}[t]
    \centering

    \begin{minipage}[t]{0.5\columnwidth}
        \vspace{0pt}
        \captionof{table}{Ablation studies on different grounding modes.}
        \label{tab:reasoning_paradigm_ablation}
        \renewcommand{\arraystretch}{1.50}
        \resizebox{\linewidth}{!}{
        \begin{tabular}{lcccccc}
        \toprule
        \multirow{2}{*}{\makecell{\textbf{Training}\\\textbf{Strategy}}} & 
        \multirow{2}{*}{\textbf{COCO}} & 
        \multirow{2}{*}{\textbf{Ref-L4}} & 
        \multicolumn{3}{c}{\textbf{PR-Bench}} & 
        \multirow{2}{*}{\textbf{AVG}} \\
        \cmidrule(lr){4-6}
        &&& mAcc$_{API}$ & mAcc$_{RC}$ & N-Acc & \\
        \midrule
       \makecell{Perceptual\\Only} & 46.2 & 88.2 & 69.1 & 55.9 & 46.7 & 61.2 \\
        \makecell{Reason\\Only} & 43.6 & 90.4 & 67.7 & 60.4 & 45.8 & 61.6 \\
        \makecell{Adaptive\\Grounding} & 46.6 & 90.7 & 73.2 & 63.2 & 46.3 & 64.0 \\
        \bottomrule
        \end{tabular}
        }
    \end{minipage}
    \hfill
    \begin{minipage}[t]{0.48\columnwidth}
        \vspace{0pt}

        \includegraphics[width=\linewidth]{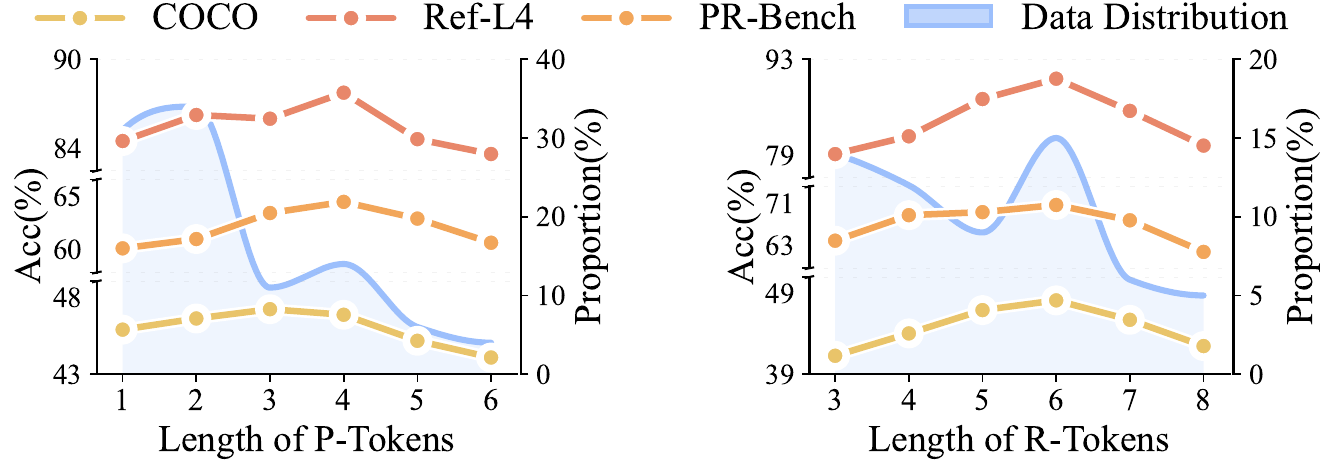}
        \vspace{-17pt}
        \captionof{figure}{
        Performance under different Token Length  Limits of (a) P-Tokens and (b) R-Tokens, together with the distribution of actual token length in training data.
        }
        \label{fig:token_length}

    \end{minipage}
\end{figure}

\noindent\textbf{Adaptive Grounding Mode.}
Table~\ref{tab:reasoning_paradigm_ablation} compares three settings: Perceptual-Only, Reason-Only, and adaptive grounding (Motto). 
The results show that the adaptive grounding mode achieves superior performance across all metrics. 
While Reason-Only excels on reasoning-based tasks such as Ref-L4 and the PR-Bench mAcc$_{RC}$ metric due to its complete inference chain, Perceptual-Only outperforms on visual benchmarks like COCO and the PR-Bench mAcc$_{API}$ and N-Acc metrics.
These observations highlight the necessity of direct grounding without extra reasoning.
Applying complex reasoning to simple queries inevitably deviates from image-centric perception, resulting in grounding hallucinations. 
Therefore, Motto dynamically selects the optimal mode based on aggregated context, effectively balancing precise perception with complex reasoning.

\noindent\textbf{Token Length Limits and Distribution Analysis.}
Figure~\ref{fig:token_length} illustrates the impact of varying maximum token limits for P-Tokens ($K_{\max}^{P}$) and R-Tokens ($K_{\max}^{R}$). 
Benefiting from explicit spatial alignment, Motto maintains stable performance across different token limits, peaking at $K_{\max}^{P}=4$ and $K_{\max}^{R}=6$. 
Notably, the distribution of actual token lengths in training data also aligns with these peaks. 
Therefore, although a few multi-object scenarios require longer sequences, setting a higher limit to cover these rare cases increases training instability and reduces performance. 
Consequently, we adopt $K_{\max}^{P}=4$ and $K_{\max}^{R}=6$ as the optimal configuration.

\subsection{Visualizations}
Figure~\ref{fig:visualize} visualizes the heatmaps of different token designs. For a fair comparison, we evaluate perceptual and reasoning abilities using the same number of tokens.
As shown in the top two rows, Motto operates in perceptual mode, where each P-token acts as a spatial anchor for a distinct region. In contrast, the heatmaps of \verb|<Det>| tokens exhibit a homogeneous distribution. Due to the lack of inherent spatial references, \verb|<Det>| tokens often fail to capture potential objects during decoding.
In the bottom two rows, we present scenarios that require reasoning, where \verb|<Det>| tokens misidentify the target from the beginning, leading to an incorrect final result. 
In contrast, Motto operates in reasoning mode: P-tokens first perform comprehensive perception, followed by R-tokens focusing on the target through a reasoning chain, which leads to robust localization.

\begin{figure}
  \includegraphics[width=1\linewidth]{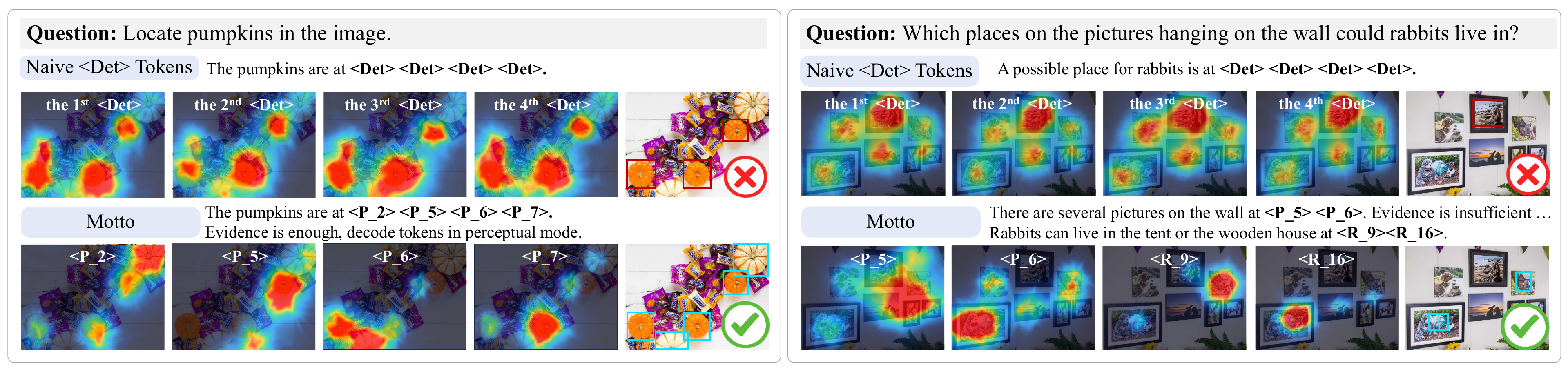}
  \caption{Comparison of heatmaps between the perceptual and reasoning modes. The visualizations of baselines using naive special tokens <Det> are also included.}
  \label{fig:visualize}
\end{figure}

\section{Conclusion}


We propose Motto, a unified grounding framework that bridges the gap between fine-grained perception and complex reasoning. By integrating Spatially-Grounded Thought Tokenization, Motto achieves precise localization with explicit visual interpretability. Furthermore, our Context-Adaptive Chain-of-Tokens enables the model to dynamically balance perceptual and reasoning modes for diverse queries. Extensive evaluations including the proposed PR-Bench show that Motto achieves state-of-the-art performance, establishing a strong baseline for free-form multimodal grounding.


\newpage
\bibliographystyle{plainnat}
\bibliography{paper}

\appendix
\clearpage
\setcounter{page}{1}
\section{In-Depth Analysis}
\subsection{Fine-Grained Adaptive Grounding Modes}
We report the mode selection tendencies of Motto across different benchmarks and their subcategories in Fig~\ref{fig:mode_proportion}. The results show that mode selection is strongly correlated with task complexity. 
In detection or short phrase grounding benchmarks like COCO and RefCOCO/+/g, the model predominantly uses perceptual mode for direct grounding, while in Ref-L4, which focuses on long-caption grounding, reasoning mode is more frequently used.
This trend is particularly evident in PR-Bench, where perceptual mode dominates in Visual-Cue Perception tasks, while reasoning mode takes the lead in Compositional Reasoning tasks, highlighting the shift towards more complex, semantically aligned tasks.
 Additionally, reasoning mode is frequently used for cross-image reasoning in MIG-Bench.

\begin{figure}[!htbp]
 \centering
  \includegraphics[width=0.7\linewidth]{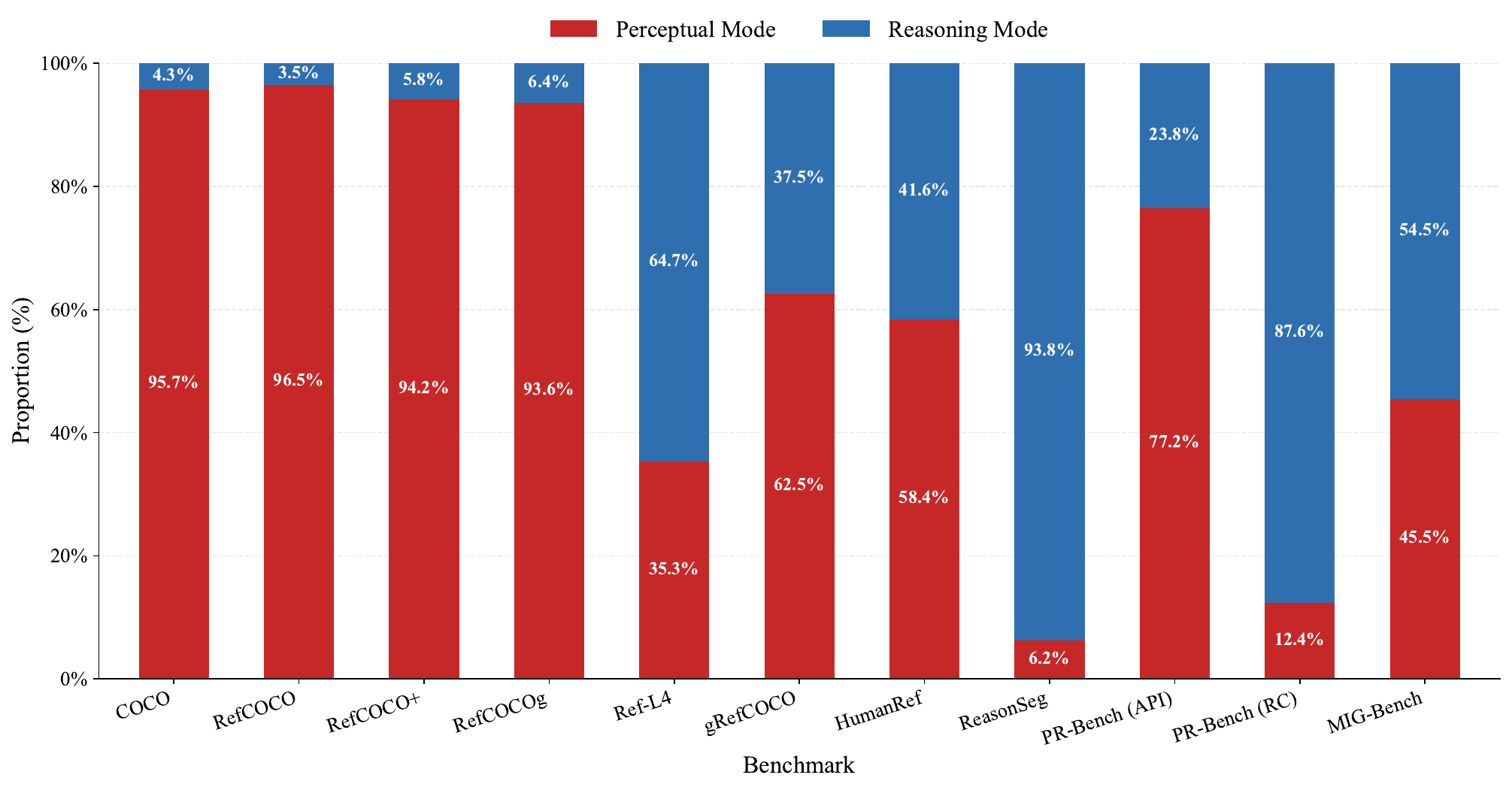}
  \caption{Proportion of perceptual mode and reasoning mode across different benchmarks.}
  \label{fig:mode_proportion}
\end{figure}

\subsection{Efficiency Analysis}
As shown in Table~\ref{tab:horizontal_efficiency}, we evaluate the inference efficiency of Motto  on single-object, multi-object, and multi-image tasks by randomly sampling 200 instances per task to analyze the end-to-end average wall-clock time (Avg. Wall Time), output sequence length (Avg. Seq Length), and grounding accuracy (Acc). 
The single-object setting is built from RefCOCO and Ref-L4, the multi-object setting is built from COCO and HumanRef, and the multi-image setting is built from MIG-Bench. The multi-object setting contains 5.1 target objects per query on average.
Methods based on direct coordinate prediction (e.g., Qwen3-VL-2B, Rex-Omni-3B) exhibit latency primarily constrained by sequence length; furthermore, they suffer from significant performance degradation in multi-object task due to error propagation. Detect-and-retrieval methods, exemplified by VLM-FO1, incurs substantial computational overhead due to the preliminary stage of generating region proposals via an external model. 
In contrast, Motto demonstrates superior adaptive capability by dynamically switching between perceptual and reasoning modes.

\section{Training Data Details}
\subsection{Training Data Composition}
Table~\ref{tab:training_data_composition} summarizes the training data used in our two-stage training pipeline.
Stage~1 uses Object365 for pre-training, and Stage~2 uses a mixed fine-tuning set covering diverse grounding tasks.
For Stage~2, we further organize the data into perceptual and reasoning subsets.
For Stage~2, we further organize the data into perceptual and reasoning data, as shown in Figure~\ref{fig:stage2_data_composition}.

\begin{table}[!htbp] 
\centering
\caption{Efficiency analysis of Motto compared with representative models. We report average wall-clock time (Time), sequence length (Len.), and accuracy (Acc.) across three tasks.}
\label{tab:horizontal_efficiency}
\small 
\setlength{\tabcolsep}{2.5pt} 

\begin{tabular}{l ccc ccc ccc}
\toprule
\multirow{2}{*}{\textbf{Model}} & \multicolumn{3}{c}{\textbf{Single-object}} & \multicolumn{3}{c}{\textbf{Multi-object}} & \multicolumn{3}{c}{\textbf{Multi-image}} \\
\cmidrule(lr){2-4} \cmidrule(lr){5-7} \cmidrule(lr){8-10}
& Time(s) & Len. & Acc. & Time(s) & Len. & Acc. & Time(s) & Len. & Acc. \\
\midrule
\multicolumn{10}{l}{\textbf{\textit{Perceptual Mode Samples}}} \\
Qwen3-VL-2B~\cite{bai2025qwen3} & 1.5 & 44.6 & 83.0 & 2.4 & 74.5 & 42.6 & 1.3 & 35.7 & 25.0 \\
Rex-Omni~\cite{jiang2025detect}   & 0.6 & 16.0 & 84.5 & 1.8 & 46.7 & 49.8 & 1.2 & 28.8 & 14.0 \\
VLM-FO1~\cite{liu2025vlm}    & 1.1 & 13.9 & 86.5 & 1.4 & 18.2 & 47.6 & 1.6 & 20.9 & 11.5 \\
 \rowcolor{gray!10} \textbf{Motto} & \textbf{0.9} & \textbf{23.2} & \textbf{92.9} & \textbf{1.2} & \textbf{28.6} & \textbf{62.1} & \textbf{1.0} & \textbf{26.3} & \textbf{76.2} \\
\midrule
\multicolumn{10}{l}{\textbf{\textit{Reasoning Mode Samples}}} \\
Qwen3-VL-2B~\cite{bai2025qwen3} & 2.1 & 68.6 & 68.0 & 2.9 & 106.7 & 24.3 & 2.3 & 80.1 & 17.5 \\
Rex-Omni~\cite{jiang2025detect}   & 1.9 & 47.5 & 43.5 & 2.5 & 69.9 & 31.2 & 2.0 & 45.9 & 9.5  \\
VLM-FO1~\cite{liu2025vlm}    & 1.6 & 19.4 & 33.0 & 2.0 & 23.9 & 38.3 & 1.8 & 21.2 & 12.0 \\
\rowcolor{gray!10} \textbf{Motto} &\textbf{1.8} & \textbf{40.5} & \textbf{72.4} & \textbf{2.0} & \textbf{45.9} & \textbf{60.2} & \textbf{1.9} & \textbf{44.2} & \textbf{62.1} \\
\bottomrule
\end{tabular} 
\end{table}



\begin{figure}[t]
    \centering

    \begin{minipage}[t]{0.5\columnwidth}
        \vspace{0pt}
        \captionof{table}{Training data composition of Motto. REC denotes referring expression comprehension.}
        \label{tab:training_data_composition}
        \renewcommand{\arraystretch}{1.22}
        \resizebox{\linewidth}{!}{
        \begin{tabular}{l c c}
        \toprule
        \textbf{Task} & \centering\textbf{Datasets} & \textbf{\# Samples} \\
        \midrule
        
        \multicolumn{3}{c}{\textbf{Stage 1: Pre-training}} \\
        Object Detection & Object365 & 7M \\
        \midrule
        
        \multicolumn{3}{c}{\textbf{Stage 2: Fine-tuning}} \\
        Object Detection & Object365 & 700K \\
        
        REC & VisualGenome & 3M \\
        REC & RefCOCO & 120K \\
        REC & RefCOCO+ & 120K \\
        REC & RefCOCOg & 80K \\
        REC & Rexverse2M & 1M \\
        
        General REC & gRefCOCO & 190K \\
        General REC & HumanRef & 45K \\
        General REC & ReasonSeg & 1K \\
        
        Omni Referring & OmniRef & 110K \\
        Multi-image Grounding & MGrounding-630K & 630K \\
        
        \bottomrule
        \end{tabular}
        }
    \end{minipage}
    \hfill
    \begin{minipage}[t]{0.48\columnwidth}
        \vspace{0pt}

        \includegraphics[width=\linewidth]{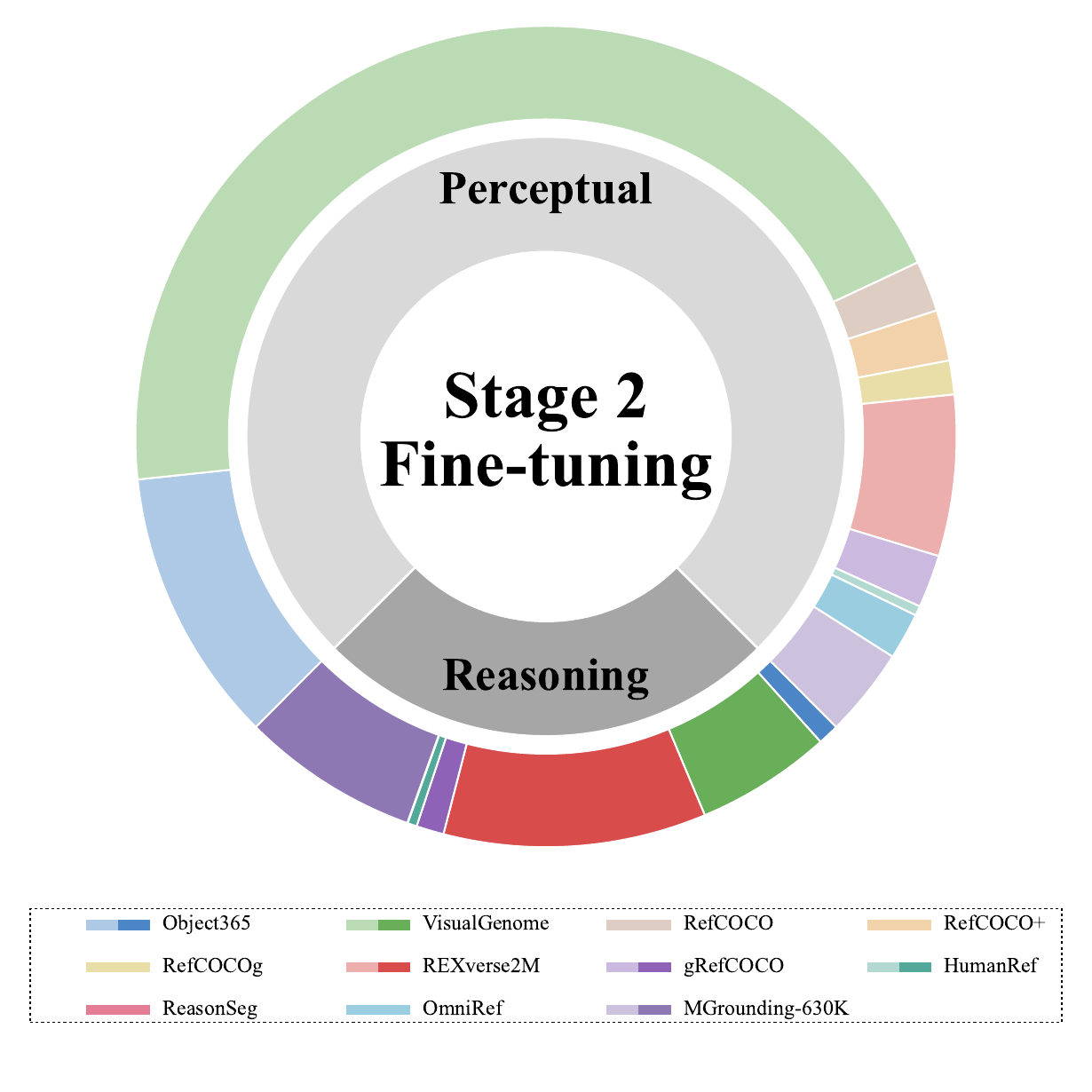}
        \vspace{-30pt}
        \caption{Distribution of perceptual and reasoning data in Stage~2 fine-tuning.}
        \label{fig:stage2_data_composition}

    \end{minipage}
\end{figure}

\subsection{Data Construction Prompts}
Figure~\ref{fig:data_construction_prompts} presents the prompt templates used in our referring grounding data construction pipeline.
The three prompts are designed for direct grounding probing, evidence diagnosis, and error reflection, respectively.
They provide the intermediate annotations used to construct perception data and reasoning data.

\begin{figure*}[!htbp]
  \includegraphics[width=\linewidth]{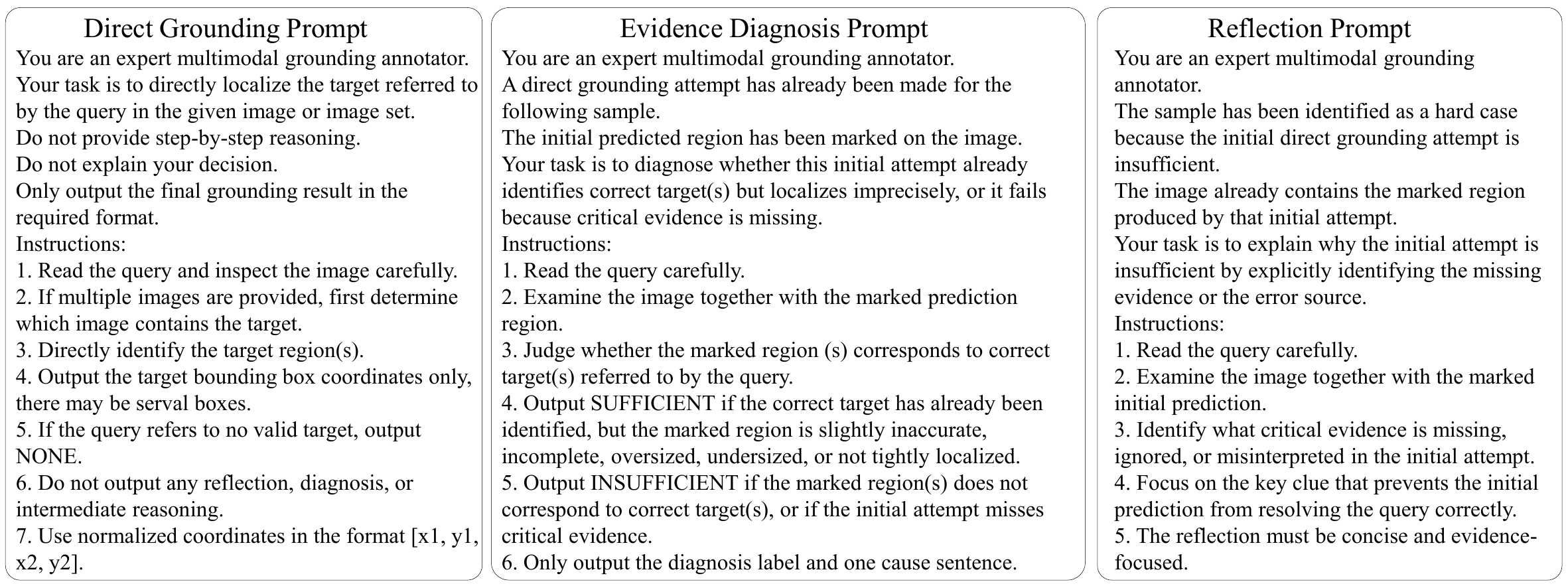}

  \caption{Prompt templates used in referring grounding data construction.}
  \label{fig:data_construction_prompts}
\end{figure*}

\section{Additional Main Results}

\subsection{Main Results on Additional Backbones}
Table~\ref{tab:full_comparison_results} reports the main results of Motto on additional backbones including Qwen2.5-VL-3B and Qwen3-VL-4B.
The results show that Motto consistently improves grounding performance across different backbones and different model scales, indicating that the proposed framework is not tied to a specific backbone.
\begin{table*}[h]
\centering
\caption{Performance comparison of different models across multiple grounding benchmarks.}
\label{tab:full_comparison_results}
\resizebox{\textwidth}{!}{%
\begin{tabular}{lccccccccccccc}
\toprule
\multirow{2}{*}{\textbf{Model}} & \textbf{RefCOCO} & \textbf{Ref-L4} & \multicolumn{2}{c}{\textbf{gRefCOCO}} & \multicolumn{2}{c}{\textbf{HumanRef}} & \textbf{Reason} & \multicolumn{2}{c}{\textbf{PR-Bench}} & \textbf{MIG-Bench} & \multicolumn{3}{c}{\textbf{COCO (mAP)}} \\
& \textbf{Avg.} & \textbf{Acc} & \textbf{Pr} & \textbf{N-Acc} & \textbf{DF1} & \textbf{Rej} & \textbf{Seg} & \textbf{mAcc} & \textbf{N-Acc} & \textbf{AVE} & \textbf{Text} & \textbf{Vis-G} & \textbf{Vis-I} \\
\midrule
\textbf{\textit{General MLLMs}} & & & & & & & & & & & & & \\
Qwen3-VL-2B~\cite{bai2025qwen3} & 83.5 & 85.4 & 46.5 & 25.1 & 58.9 & -- & 49.8 & 52.8 & -- & 21.4 & 34.9 & 11.6 & 25.3 \\
Youtu-VL-4B~\cite{wei2026youtu} & 91.9 & 83.3 & 25.7 & -- & 55.3 & -- & 51.3 & 53.2 & -- & 32.9 & 47.1 & 18.4 & 21.0 \\
InternVL3.5-8B~\cite{wang2025internvl3} & 89.7 & 80.9 & 31.2 & -- & 40.0 & 33.1 & 38.3 & 41.5 & -- & -- & -- & -- & -- \\
Qwen3-VL-8B~\cite{bai2025qwen3} & 87.9 & 88.5 & 55.4 & 55.0 & 64.9 & 47.9 & 56.4 & 59.4 & 16.2 & 31.8 & 36.8 & 20.8 & 29.5\\
GLM-4.1V-Thinking-9B~\cite{hong2025glm} & 86.0 & 84.8 & 49.3 & 29.0 & 58.6 & 34.5 & 57.6 & 57.9 & -- & 46.7 & 27.2 & 14.5 & 10.4 \\
Qwen3.5-9B~\cite{qwen3.5} & 88.1 & 89.0 & 56.3 & -- & 72.0 & 10.8 & 56.6 & 63.7 & 11.9 & 31.7 & 37.7 & 20.8 & 32.6 \\
Qwen3.5-27B~\cite{qwen3.5} & 90.9 & 90.2 & 56.7 & -- & 71.5 & 13.4 & 65.7 & 65.3 & -- & 42.7 & 38.8 & 22.2 & 33.3 \\
InternVL3.5-38B~\cite{wang2025internvl3} & 89.2 & 85.5 & 36.0 & -- & 51.1 & -- & 51.7 & 48.5 & 12.7 & -- & -- & -- & -- \\
\midrule
\textbf{\textit{Grounding-Specialized MLLMs}} & & & & & & & & & & & & & \\
T-Rex-2~\cite{jiang2024t} & -- & -- & -- & -- & -- & -- & -- & -- & -- & -- & 52.2 & 46.5 & 58.5 \\
VLM-R1~\cite{shen2025vlm} & 86.3 & 80.9 & 45.7 & 30.6 & 46.4 & 27.1 & 26.5 & 54.4 & -- & -- & -- & -- & -- \\
VLM-FO1~\cite{liu2025vlm} & 88.6 & 84.4 & 58.7 & 52.5 & 82.6 & 46.4 & 46.7 & 53.9 & 11.4 & -- & 44.4 & 14.1 & 11.0 \\
Rex-Omni-3B~\cite{liu2025vlm} & 83.9 & 78.0 & 38.0 & 33.6 & 74.9 & -- & 40.6 & 53.7 & -- & 11.9 & -- & 17.4 & -- \\
Chatrex~\cite{jiang2024chatrex} & 89.2 & 80.0 & 29.9 & -- & 55.6 & -- & 23.0 & 49.5 & -- & -- & -- & -- & -- \\
ROD-MLLM~\cite{yin2025rod} & 86.9 & -- & 53.5 & 57.8 & -- & -- & -- & -- & -- & -- & -- & -- & -- \\
Migician~\cite{li2025migician}  & 88.1 & 73.2 & 29.4 & 14.4 & 45.9 & 26.0 & 47.3 & 52.3 & -- & 63.8 & 10.0 & 7.1 & 9.3 \\
DeepEyes~\cite{zheng2025deepeyes} & 86.7 & 85.2 & 50.3 & 38.4 & 50.3 & 22.8 & 68.6 & 45.7 & -- & -- & -- & -- & -- \\
VGent~\cite{kang2025vgent} & 90.2 & -- & -- & -- & -- & -- & -- & -- & -- & -- & -- & -- & -- \\
\rowcolor{gray!10} \textbf{Motto{\scriptsize{Qwen3-VL-2B}}} & \textbf{92.6} & \textbf{92.5} & \textbf{69.6} & \textbf{58.0} & \textbf{83.0} & \textbf{53.4} & \textbf{71.2} & \textbf{71.7} & \textbf{46.9} & \textbf{74.5} & \textbf{56.4} & \textbf{51.7} & \textbf{58.9} \\
\rowcolor{gray!10} \textbf{Motto{\scriptsize{Qwen3-VL-4B}}} & \textbf{94.8} & \textbf{94.3} & \textbf{73.3} & \textbf{58.3} & \textbf{84.1} & \textbf{54.3} & \textbf{74.8} & \textbf{74.1} & \textbf{47.5} & \textbf{76.6} & \textbf{57.7} & \textbf{53.1} & \textbf{60.4} \\
\rowcolor{gray!10} \textbf{Motto{\scriptsize{Qwen2.5-VL-3B}}} & \textbf{92.1} & \textbf{89.4} & \textbf{67.4} & \textbf{57.6} & \textbf{82.7} & \textbf{53.5} & \textbf{70.7} & \textbf{68.1} & \textbf{47.4} & \textbf{72.8} & \textbf{55.0} & \textbf{50.7} & \textbf{57.5} \\
\bottomrule
\end{tabular}%
}
\end{table*}

\begin{table*}[h]
\centering

\caption{Performance comparison of different models on MIG-Bench~\cite{li2025migician}. OT, MV, GG, and Co-Re respectively mean object tracking,multi-view grounding, group grounding, and correspondence.}
  \resizebox{\textwidth}{!}{\begin{tabular}{lccccccccccc}
\toprule

\multirow{3}{*}{\textbf{Model}}& \multicolumn{3}{c}{\textbf{Spontaneous Grounding}} 
& \multicolumn{7}{c}{\textbf{Referential Grounding}} &\multirow{3}{*}{\textbf{AVE}} \\
\cmidrule(lr){2-4} \cmidrule(lr){5-11}

 & \multicolumn{2}{c}{\textbf{Difference} }
& \textbf{Similarity}
& \multicolumn{4}{c}{\textbf{Visual Reference}} 
& \textbf{Textual }
& \multicolumn{2}{c}{\textbf{Visual+Textual}} \\
\cmidrule(lr){2-3}\cmidrule(lr){4-4} \cmidrule(lr){5-8}\cmidrule(lr){9-9} \cmidrule(lr){10-11}

 & \textbf{Static} & \textbf{Robust} & \textbf{Common }
& \textbf{OT} & \textbf{MV} & \textbf{Region} & \textbf{Refer} 
& \textbf{GG }
& \textbf{Reason} & \textbf{Co-Re} & \\

\midrule

InternVL3.5-8B~\cite{wang2025internvl3} & 2.5 & 3.2 & 13.6 & 10.2 & 6.3 & 2.0 & 12.1 & 17.6 & 5.0 & 0.9 & 7.3 \\
 Qwen3-VL-2B~\cite{bai2025qwen3} & 7.4 & 4.5 & 30.3 & 12.5 & 6.3 & 3.7 & 25.3 & 20.8 & 10.9 & 4.3 & 12.6 \\
Qwen3-VL-8B~\cite{bai2025qwen3} & 40.7 & 24.5 & 38.8 & 30.0 & 22.9 & 26.1 & 23.2 & 53.9 & 32.7 & 25.6 &31.8\\
Youtu-VL-4B~\cite{wei2026youtu} & 21.8 & 31.9 & 46.1 & 49.6 & 38.5 & 0.3 & 60.6 & 37.9 & 27.7 & 14.5 & 32.9 \\
Qwen3.5-9B~\cite{qwen3.5} & 53.6 & 11.7 & 54.9 & 29.5 & 30.2 & 29.5 & 30.3 & 44.9 & 24.8 & 7.7 & 31.7 \\
GLM-4.1V-Thinking-9B~\cite{hong2025glm} & 44.5 & 50.0 & 55.9 & 28.1 & 47.7 & 6.5 & 75.8 & 63.5 & 63.4 & 31.6 & 46.7 \\
Qwen3.5-27B~\cite{qwen3.5} & 56.3 & 21.3 & 68.5 & 28.0 & 37.9 & 33.9 & 51.5 & 66.1 & 35.6 & 28.2 & 42.7 \\
Migician~\cite{li2025migician} & 65.2 & 46.8 & 84.2 & 70.7 & 60.1 & 74.3 & 76.8 & 66.5 & 59.4 & 34.2 & 63.8 \\
\rowcolor{gray!10} \textbf{Motto{\scriptsize{Qwen3-VL-2B}}} &\textbf{69.5} & \textbf{56.4} & \textbf{87.0} & \textbf{79.6} & \textbf{66.3} & \textbf{87.9} & \textbf{90.9} & \textbf{85.4} & \textbf{67.3} & \textbf{54.7} & \textbf{74.5}\\
\rowcolor{gray!10}  \textbf{Motto{\scriptsize{Qwen3-VL-4B}}} &\textbf{69.7} & \textbf{60.6} & \textbf{90.7} & \textbf{82.4} & \textbf{68.4} & \textbf{88.1} & \textbf{91.9} & \textbf{88.2} & \textbf{70.3} & \textbf{55.6} & \textbf{76.6}\\
\rowcolor{gray!10}  \textbf{Motto{\scriptsize{Qwen2.5-VL-3B}}} &\textbf{68.2} & \textbf{52.1} & \textbf{86.5} & \textbf{77.1} & \textbf{63.9} & \textbf{87.0} & \textbf{88.9} & \textbf{83.8} & \textbf{66.3} & \textbf{53.8} & \textbf{72.8}  \\
\bottomrule
\end{tabular}}
\label{tab:suppl_full_migbench}
\end{table*}

\subsection{Detailed Results on MIG-Bench}
Table~\ref{tab:suppl_full_migbench} presents the fine-grained results on multi-image grounding benchmarks. Motto achieves state-of-the-art performance across all tasks on MIG-Bench.

\begin{table*}[h]
\centering
\caption{Fine-grained evaluation on PR-Bench across six evaluation aspects. }
\resizebox{1.0\linewidth}{!}{
\begin{tabular}{l c ccc ccc ccc}
\toprule
\multirow{2}{*}{\textbf{Models}}
& \multirow{2}{*}{\textbf{Size}}
& \multicolumn{3}{c}{\textbf{Overall}}
& \multicolumn{3}{c}{\textbf{Visual-cue Perception}}
& \multicolumn{3}{c}{\textbf{Compositional Reasoning}} \\
\cmidrule(lr){3-5} \cmidrule(lr){6-8} \cmidrule(lr){9-11}
& & \textbf{$mAcc$}& mAcc$_{API}$ & mAcc$_{RC}$ & \textbf{Attribute} & \textbf{Position} & \textbf{Interaction} & \textbf{Relation} & \textbf{Commonsense} & \textbf{Rejection} \\

\midrule
\multicolumn{11}{l}{\textbf{\textit{General MLLMs}}} \\
Qwen3-VL~\cite{bai2025qwen3} &2B & 52.8 &55.7 & 48.2& 59.5 & 58.0 & 50.0 & 46.7 & 49.6 & --\\
Youtu-VL~\cite{wei2026youtu}&4B & 53.2&56.7& 47.9& 61.5 & 57.2 & 51.4 & 49.2 & 46.6 &--\\
InternVL3.5~\cite{wang2025internvl3}& 8B & 41.5&44.1&37.6 & 45.7 & 41.2 & 45.3 & 37.8 & 37.3 & -- \\
Qwen3-VL~\cite{bai2025qwen3}& 8B & 59.4 &62.0&55.6& 64.8 & 62.6 & 58.7 & 55.8 & 55.4 & 16.2 \\
GLM-4.1V-Thinking~\cite{hong2025glm} & 9B &57.9 &61.2&52.9&63.6 &61.5 &58.6 &52.4 &53.4 &--  \\
Qwen3.5~\cite{qwen3.5}&9B &63.7 &66.3&59.7& 68.9 & 66.9 & 63.1 & 59.4 & 60.0 & 11.9\\
Qwen3.5~\cite{qwen3.5}&27B & 65.3 &67.6&61.9& 68.4 & 67.3 & 67.0 & 62.7 & 61.0 &--\\
InternVL3.5~\cite{wang2025internvl3} &38B & 48.5 &51.7&43.7& 52.0 & 50.8 & 52.3 & 43.9 & 43.4 & 12.7  \\

\midrule

\multicolumn{11}{l}{\textbf{\textit{Grounding-Specialized MLLMs}}} \\
VLM-R1~\cite{shen2025vlm}& 3B & 54.4 &57.0&50.5& 59.0 & 58.0 & 54.1 & 47.8 & 53.2 & --  \\
VLM-FO1~\cite{liu2025vlm} &3B & 53.9 &57.3&48.8& 57.9 & 56.8 & 57.2 & 49.6 & 47.9 & 11.4\\
Rex-Omni~\cite{jiang2025detect} &3B & 53.7 &58.8&46.0& 59.5 & 58.0 & 58.9 & 46.8 & 45.1 &--\\
DeepEyes~\cite{zheng2025deepeyes} &7B & 45.7 &47.3&43.1& 49.7 & 46.3 & 46.0 & 42.6 & 43.6 & -- \\
ChatRex~\cite{jiang2024chatrex}&  7B  & 49.5&53.0&44.3 & 54.7 & 51.1 & 53.2 & 45.1 & 43.4 & -- \\
Migician~\cite{li2025migician}& 7B & 52.3 &56.6&45.8& 57.3 & 59.7 & 52.8 & 45.4 & 46.1 & -- \\
LocateAnything~\cite{wang2025locateanything} &3B &57.4 &61.7 & 50.8& 63.1 & 61.4 &60.8 &50.0&51.6& --  \\

\rowcolor{gray!10} \textbf{Motto{\scriptsize{Qwen3-VL-2B}}} & 2B & \textbf{71.7} &\textbf{74.4}&\textbf{67.6}& \textbf{75.8} & \textbf{74.5}& \textbf{72.9} & \textbf{66.3} & \textbf{68.9} &\textbf{46.9} \\
\rowcolor{gray!10} \textbf{Motto{\scriptsize{Qwen3-VL-4B}}} & 4B & \textbf{74.1} &\textbf{76.2}&\textbf{71.0}& \textbf{78.4} & \textbf{76.2} & \textbf{74.0} & \textbf{69.8} & \textbf{72.1}& \textbf{47.5} \\
\rowcolor{gray!10} \textbf{Motto{\scriptsize{Qwen2.5-VL-3B}}} & 3B & \textbf{68.1} & \textbf{69.9}&\textbf{65.4}&\textbf{72.6} & \textbf{69.4} & \textbf{67.8} & \textbf{64.2} & \textbf{66.5} & \textbf{47.4} \\
\bottomrule
\end{tabular}}
\label{tab:suppl_full_prbench}
\end{table*}
\subsection{Detailed Results on PR-Bench}
Table~\ref{tab:suppl_full_prbench} presents the detailed results on PR-Bench. Motto achieves state-of-the-art performance across all evaluation dimensions, outperforming both specialist and generalist models.

\section{Additional Ablation Studies}
We conduct additional ablation studies on a 3M sample subset, adopting the same hyperparameters as the full-dataset training.

\subsection{Ablation on Training Strategy}

Table~\ref{tab:ablation_train_stage} compares our two-stage training strategy with a single-stage baseline. 
The two-stage approach achieves a 1.1\% average improvement, with significant gains in reasoning-centric tasks such as Ref-L4 and PR-Bench mAcc$_{RC}$. 
By aligning perceptual scanning in Stage 1 and introducing the full Scan-Focus-Action paradigm in Stage 2, we establish a more coherent training process across tasks and maximize the model's capability for complex queries.

\subsection{Ablation on $L_{cls}$}

Table~\ref{tab:l_cls_ablation} evaluates the top-$K$ hard negative selection in $\mathcal{L}_{cls}$. Without selection, performance is limited by excessive negative sample interference. The setting of $K=20$ yields optimal results, particularly achieving peak performance on multi-target benchmarks like COCO and gRefCOCO. While an insufficiently small $K$ fails to provide enough contrastive signals, an excessively large $K$ leads to performance drops as easy negatives dominate the training process. 
Consistent gains observed across PR-Bench subcategories further confirm that balanced selection is essential for stable and effective classification optimization.

\begin{table}[!htbp]
    \centering

    \begin{minipage}[t]{0.48\textwidth}
        \vspace{0pt}
        \caption{Ablation studies on training strategies.}
        \label{tab:ablation_train_stage}

        \scriptsize
        \setlength{\tabcolsep}{4pt}
        \renewcommand{\arraystretch}{1.60}
        \resizebox{\linewidth}{!}{
        \begin{tabular}{lcccccc}
        \toprule
        \multirow{2}{*}{\textbf{Strategy}} & \multirow{2}{*}{\textbf{COCO}} & \multirow{2}{*}{\textbf{Ref-L4}} & \multicolumn{3}{c}{\textbf{PR-Bench}} & \multirow{2}{*}{\textbf{AVG}} \\
        \cmidrule(lr){4-6}
        &&& mAcc$_{API}$ & mAcc$_{RC}$ & N-Acc & \\
        \midrule
        Single-stage Training & 46.1&89.0 & 72.2 & 60.6 & 46.4 &  62.9 \\
        \midrule
        Two-stage Training\\
        \multicolumn{1}{c}{Stage 1} & 45.7 & -- & -- & -- & -- & -- \\
        \multicolumn{1}{c}{Stage 2} &46.6 &90.7 & 73.2 & 63.2 & 46.3  & 64.0 \\
        
        \bottomrule
        \end{tabular}}
    \end{minipage}
    \hfill
    \begin{minipage}[t]{0.5\textwidth}
        \vspace{0pt}
        \caption{Ablation studies on top-$K$ hard negative selection in $\mathcal{L}_{cls}$, where w/o Selection corresponds to the standard $L_{cls}$ without top-$K$ hard negative selection.}
        \label{tab:l_cls_ablation}

        \scriptsize
        \setlength{\tabcolsep}{4pt}
        \renewcommand{\arraystretch}{1.02}
        \resizebox{\linewidth}{!}{
        \begin{tabular}{l c c ccc c}
        \toprule
        \multirow{2}{*}{\textbf{Setting}} & \multirow{2}{*}{\textbf{COCO}} & \textbf{gRefCOCO} & \multicolumn{3}{c}{\textbf{PR-Bench}} & \multirow{2}{*}{\textbf{AVG}} \\
        \cmidrule(lr){4-6}
         && Pr & mAcc$_{API}$ & mAcc$_{RC}$ & N-Acc& \\
        \midrule
        w/o Selection & 45.1 & 61.9 & 72.4 & 61.7 & 46.6 & 57.5 \\
        $K=10$ & 44.8 & 63.3 & 72.2 & 62.8 & 46.2 & 57.8 \\
         $K=20$ & 46.6 & 64.2 & 73.2 & 63.2 & 46.3 & 58.7 \\
        $K=30$ & 45.8 & 63.7 & 73.0 & 62.4 & 46.4 & 58.2 \\
        \bottomrule
        \end{tabular}}
    \end{minipage}
\end{table}

\subsection{Ablation on Switch Adapter}

Table~\ref{tab:switch_adapter_ablation} evaluates different settings of Switch Adapter. Relying solely on text embeddings or image features yields suboptimal results, as it fails to capture complementary cross-modal cues. Our full Switch Adapter achieves the best average performance by aggregating cross-modal context.

\subsection{Ablation on Focus Adapter}

Table~\ref{tab:focus_adapter_detailed} evaluates different settings of Focus Adapter. Starting from the baseline configuration integrating P+R-Tokens, the Switch Adapter, and Textual Reflection, we investigate how different configurations of visual evidence aggregation impact grounding performance. Results show that relying on either global or target-focused features in isolation yields limited improvements, as they either lack spatial resolution or fail to capture broader scene context. 
The full Focus Adapter achieves the best performance by hierarchically aggregating both global and target-focused evidence, particularly in reasoning-centric tasks such as PR-Bench RC and MIG-Bench.

\begin{table}[!htbp]
    \centering

    \begin{minipage}[t]{0.48\textwidth}
        \vspace{0pt}
        \caption{Ablation studies of Switch Adapter configurations. Full Switch Adapter aggregates cross-modal context from both text embeddings and visual features.}
        \label{tab:switch_adapter_ablation}

        \scriptsize
        \setlength{\tabcolsep}{4pt}
        \renewcommand{\arraystretch}{1.35}
        \resizebox{\linewidth}{!}{
        \begin{tabular}{lcccccc}
        \toprule
        \multirow{2}{*}{\textbf{Setting}} & \multirow{2}{*}{\textbf{COCO}} & \multicolumn{3}{c}{\textbf{PR-Bench}} & \multirow{2}{*}{\textbf{MIG-Bench}} & \multirow{2}{*}{\textbf{AVG}} \\
         && mAcc$_{API}$ & mAcc$_{RC}$ & N-Acc&  & \\
        \midrule
        \multicolumn{7}{l}{\textbf{\textit{Baselines}}} \\
        P-Tokens only & 46.2 & 69.1 & 55.9 & 46.7 & 62.6 & 56.1 \\
        P + R-Tokens  & 45.9 & 70.3 & 59.4 & 45.8 & 65.3 & 57.3 \\
        \midrule
        \multicolumn{7}{l}{\textbf{\textit{Switch Adapter Settings}}} \\
        Text embeddings only & 46.0 & 70.8 & 59.8 & 46.1 & 65.6 &  57.6\\
        Visual features only & 46.1 & 71.0 & 59.6 & 46.2 & 65.7 &  57.7\\
        Full Switch Adapter & 46.4 & 71.6 & 60.9 & 46.5 & 66.2 & 58.3 \\
        \bottomrule
        \end{tabular}}
    \end{minipage}
    \hfill
    \begin{minipage}[t]{0.5\textwidth}
        \vspace{0pt}
        \caption{Ablation studies of Focus Adapter configurations. The baseline denotes the configuration with P+R-Tokens, Switch Adapter and Textual Reflection. Full Focus Adapter aggregates both global and target-focused visual evidence.}
        \label{tab:focus_adapter_detailed}

        \scriptsize
        \setlength{\tabcolsep}{4pt}
        \renewcommand{\arraystretch}{1.30}
        \resizebox{\linewidth}{!}{
        \begin{tabular}{l c ccc c c}
        \toprule
        \multirow{2}{*}{\textbf{Setting}} & \multirow{2}{*}{\textbf{COCO}} & \multicolumn{3}{c}{\textbf{PR-Bench}} & \multirow{2}{*}{\textbf{MIG-Bench}} & \multirow{2}{*}{\textbf{AVG}} \\
        \cmidrule(lr){3-5}
         & & mAcc$_{API}$ & mAcc$_{RC}$ & N-Acc &  & \\
        \midrule
        \multicolumn{7}{l}{\textbf{\textit{Baseline}}} \\
        P + R + Switch + Refl. & 46.7 & 72.4 & 61.7 & 46.4 & 66.6 & 58.8 \\
        \midrule
        \multicolumn{7}{l}{\textbf{\textit{Focus Adapter Settings}}} \\
        Global Features only & 46.5 & 72.8 & 62.1 & 46.3 & 66.8 & 58.9 \\
        Target-focused Features only & 46.4 & 72.7 & 62.5 & 46.2 & 67.1 & 58.9 \\
        Full Focus Adapter &46.6 & 73.2 & 63.2 & 46.3 & 67.5 & 59.4 \\
        \bottomrule
        \end{tabular}}
    \end{minipage}
\end{table}

\section{Additional Details of PR-Bench}
\label{sec:supp_prbench}

\subsection{Definition of Each Subcategory in PR-Bench}
\label{sec:supp_prbench_def}

\begin{figure*}[!t]
\centering
\includegraphics[width=\linewidth]{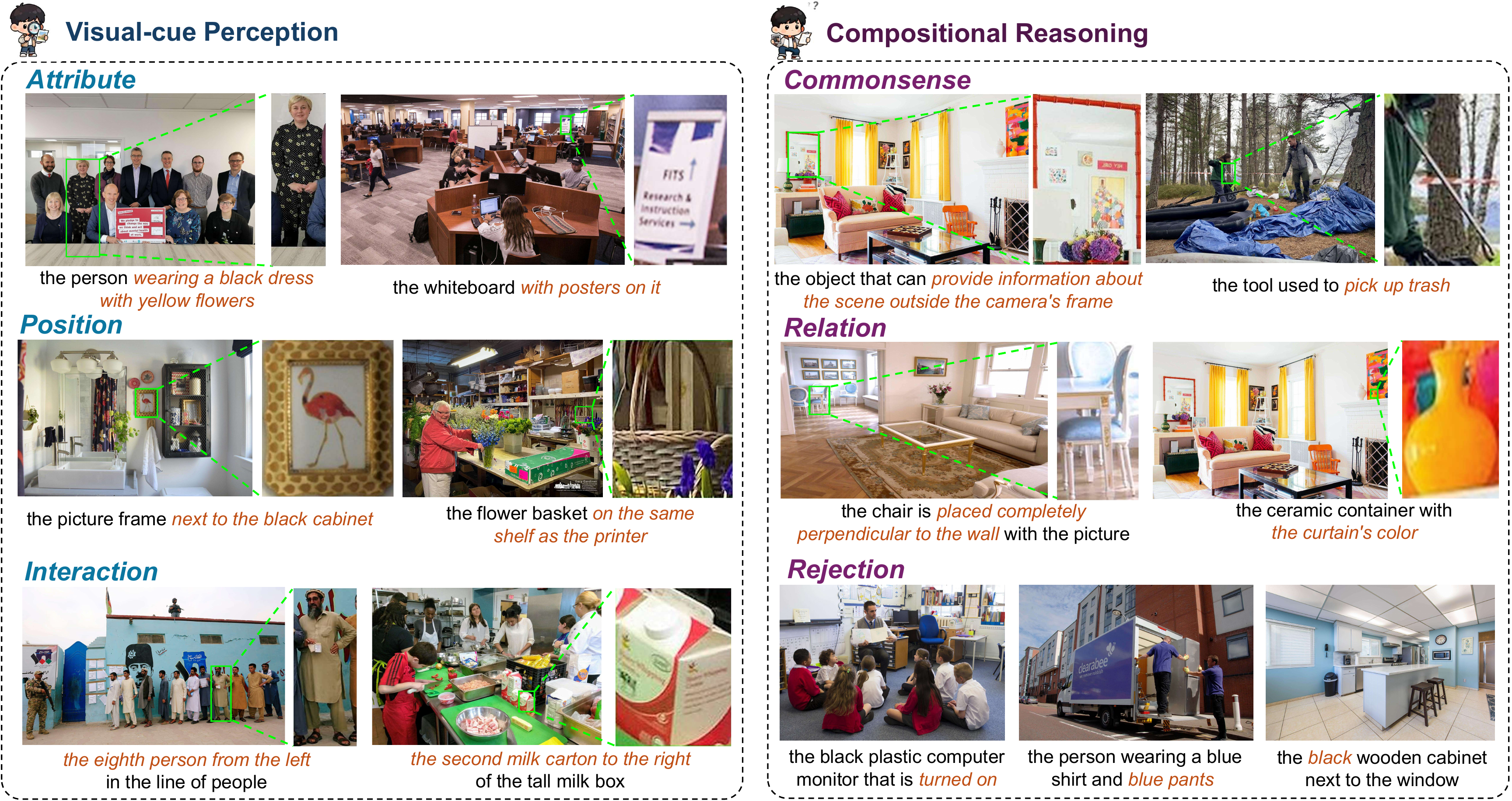}
\caption{Illustrative examples of the six PR-Bench subcategories. The three tasks on the left (Attribute, Position, and Interaction) mainly evaluate visual-cue perception, while the three on the right (Commonsense, Relation, and Rejection) require stronger compositional reasoning.}
\label{fig:supp_prbench_examples_overview}
\end{figure*}

The definitions for the six subcategories in PR-Bench are provided below, with corresponding visual examples illustrated in Fig~\ref{fig:supp_prbench_examples_overview}. These six subcategories are mutually exclusive and collectively evaluate two distinct dimensions: Visual-Cue Perception (Attribute, Position, and Interaction) and Compositional Reasoning (Relation, Commonsense, and Rejection).

\noindent{\textbf{Attribute}}. The Attribute subcategory focuses on the intrinsic and directly observable visual properties of objects, including characteristics such as color, texture, material, shape, and state.Target identification relies on distinguishing fine-grained visual cues from same-category distractors.

\noindent{\textbf{Position}}. The Position subcategory captures spatial relationships between different objects in an image. It requires localizing a target based on its relative placement to heterogeneous anchor objects in the scene.

\noindent{\textbf{Interaction}}. The Interaction subcategory focuses on the spatial arrangement and relative ordering among objects of the same category. It typically relies on ordinal or relative positioning within a homogeneous group.

\noindent{\textbf{Relation}}.The Relation subcategory identifies targets via comparative relationships with distinct anchor objects, leveraging shared or contrasting visual attributes. This category requires grounding multiple entities and then performing comparative judgment.

\noindent{\textbf{Commonsense}}. The Commonsense subcategory identifies targets through contextual or functional descriptions instead of explicit visual attributes. The referring expression therefore describes the target by its use, purpose, or context in the scene.

\noindent{\textbf{Rejection}}. The Rejection subcategory focuses on handling negation and non-existent references within referring expressions. In these cases, the referring expression contains one or more conditions that do not match any object in the image.

\subsection{Annotation Pipeline of PR-Bench}

PR-Bench is developed using images sourced from FineHARD~\cite{xie2025fg}, a large-scale dataset annotated with bounding boxes across a wide range of  resolutions. To ensure consistent quality and manageable computational demand, we filter images based on resolution variability, retaining only those between 1024$\times$1024 and 2048$\times$2048 pixels, and the distribution of Grounding DINO-detected bounding boxes. This selection process prioritizes images exhibiting multiple diverse objects and rich visual content. To generate high-quality query–bounding box  pairs across six subcategories, we propose the fine-grained annotation pipeline as depicted in Figure~\ref{fig:data_construct_pipeline}.

\begin{figure*}[htbp]
\includegraphics[width=\linewidth]{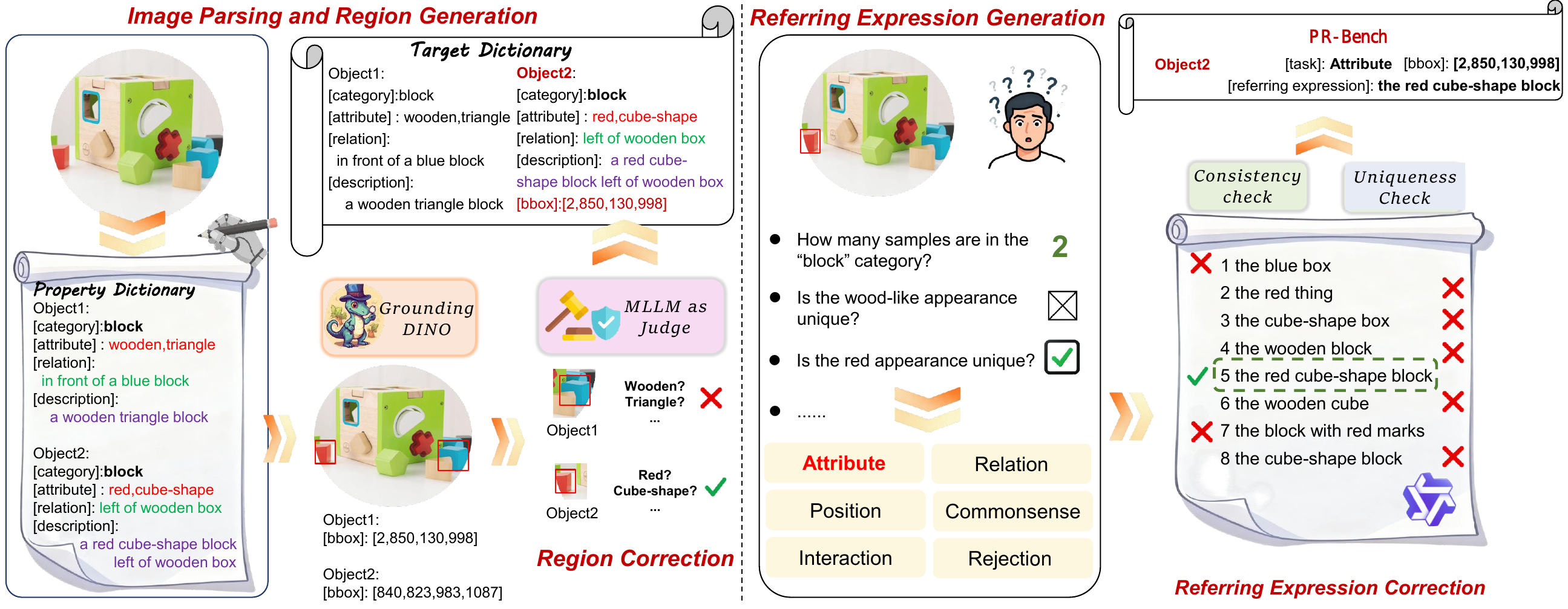}
\caption{The fine-grained referential annotation pipeline to construct PR-Bench.} 
\label{fig:data_construct_pipeline}
\end{figure*}

\noindent\textbf{Image Parsing and Region Generation.} For each image $I$, we prompt Qwen2.5-VL-72B~\cite{bai2025qwen2} to generate a structured property dictionary encoding the visual elements, including the object category $c$, inherent attributes  $a$, interactive relationships with other objects $r$, and a brief description phrase $d$. Then, we use Grounding DINO~\cite{liu2024grounding} to ground the description phrase $d$ to obtain the bounding boxes $b$. To ensure visual complexity, we retain images containing at least two distinct object categories and five  total objects. Consequently, the property for each image $I$  is structured as:
\begin{equation}
    \mathcal{E_{I}} = \{(a_j,b_j,c_j,d_j,r_j)\}_{j = 1}^{N},\, \text{with} \, |c_j| > 2, N \geq 5,
\label{Parsing}
\end{equation}
where $j$ indicates the index of specific object.

\noindent\textbf{Region Correction.} For each object $j$, we convert each element of $\mathcal{E_{I}}$  into a verification checklist. Qwen2.5-VL-72B is then prompted to evaluate whether the visual content within the target region aligns with the corresponding property in the checklist. Only objects that are fully consistent across all properties are retained, as they provide reliable  spatial coordinates and diverse property descriptions.

\noindent\textbf{Referring Expression Generation.} Once the properties are verified, we design a task selection mechanism to automatically assign each retained object to a specific task type, formalized by the selection function $\mathcal{F}_{\text{select}}$  as follows: 

\begin{equation}
T_j = \mathcal{F}_{\text{select}}(t_j, \mathcal{E}_{I}).
\end{equation}
where $t_j$  is the target object to be assigned. $\mathcal{F}_{\text{select}}$ is a human-defined rule-based selection function designed to identify challenging instances for each task by defining  specific visual conditions. For example, we assign objects that belong to the same category but exhibit distinct inherent attributes to \textit{Attribute} task, providing more fine-grained visual cues as the priors. These rules intentionally introduce visual complexity, thereby curating a set of high-quality, task-relevant seed objects. Then we prompt Qwen2.5-VL-72B to reformulate the property to generate candidate expressions for the assigned object in specific task $T_j$.

\noindent\textbf{Referring Expression Correction.} Each referring expression is validated through a two-stage verification process. First, a consistency check is adopted to verify that the visual content within the bounding box semantically aligns with the expression, filtering out hallucinated descriptions.
Second, a uniqueness check ensures that the expression unambiguously refers to a single, distinct region in the image.

\subsection{Task Selection Rules}

The detailed human-defined rule-based selection function designed to identify challenging instances for each subcategories is shown in Table~\ref{tab:supp_selection_function}.

\subsection{Statistics of PR-Bench}
PR-Bench consists of 6,000 pairs across six subcategories, as summarized in Table~\ref{tab:supp_prbench_stats}. To ensure superior data quality, we implement a rigorous human verification process following the automated pipeline. Ten expert annotators conduct a multi-round review to manually correct localization inaccuracies and refine linguistic ambiguities, retaining only the most precise and challenging samples.

\begin{table*}[!htbp]
\centering
\small
\caption{Detailed selection rules for the six PR-Bench subcategories.}
\resizebox{\textwidth}{!}{
\begin{tabular}{c l}
\toprule
\textbf{Task Category} & \textbf{Selection Rules for Target Object} \\
\midrule
Attribute &
Requires multiple same-category instances distinguished by unique intrinsic properties rather than simple naming. \\
\midrule
Position &
Localizes a target among multiple same-category instances via spatial relationships to heterogeneous anchor objects. \\
\midrule
Interaction &
Identifies the target within a group of three or more same-category instances based on ordinal position or relative arrangement. \\
\midrule
Relation &
Identifies the target via comparative reasoning against an anchor object using shared or contrasting attributes. \\
\midrule
Commonsense &
Identifies targets via functional, purposeful, or contextual descriptions rather than explicit visual attributes. \\
\midrule
Rejection &
Targets no matching object in the image, requiring the model to reject invalid references without hallucinating. \\
\bottomrule
\end{tabular}}

\label{tab:supp_selection_function}
\end{table*}
\begin{table}[!htbp]
\centering
\small
\caption{Distribution statistics of PR-Bench.}
\setlength{\tabcolsep}{4pt}
\begin{tabular}{lccccc}
\toprule
Type & Images & Phrases & Image Size & Box Size & Phrase Len. \\
\midrule
Attribute     & 701 & 1,000 & 1602$\times$1301 & 10.76\% & 9.3 \\
Position      & 735 & 1,000 & 1588$\times$1294 & 8.34\%  & 11.0 \\
Interaction   & 642 & 1,000 & 1608$\times$1292 & 9.43\%  & 10.3 \\
Relation      & 695 & 1,000 & 1619$\times$1282 & 6.15\%  & 14.4 \\
Commonsense   & 715 & 1,000 & 1627$\times$1280 & 6.14\%  & 15.5 \\
Rejection        & 764 & 1,000 & 1629$\times$1301 & N/A     & 13.8 \\
\midrule
Total         & 3102 & 6,000 & 1612$\times$1292 & 8.16\% & 12.4 \\
\bottomrule
\end{tabular}

\label{tab:supp_prbench_stats}
\end{table}

\subsection{Annotation Prompts}

We provide the prompts used in the two main annotation stages of PR-Bench construction. Figure~\ref{fig:supp_prompt_parse} presents the prompt for structured image parsing, which directs the model to decompose the scene into object-centric properties. Figure~\ref{fig:supp_prompt_expression} illustrates the prompts for task-specific referring expression generation, enforcing task-aware construction across the six subcategory definitions.

\begin{figure*}[!htbp]
\centering 
\includegraphics[width=\linewidth]{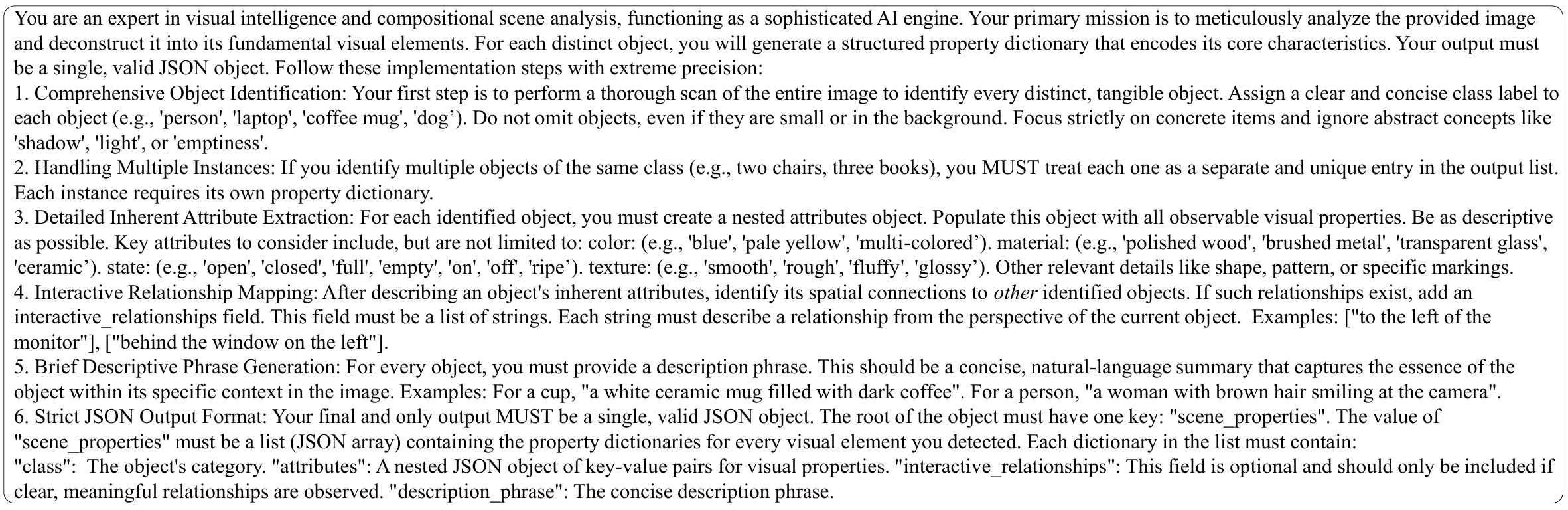}
\caption{Prompt for structured image parsing in PR-Bench, directing model to decompose scenes into object-centric properties.}
\label{fig:supp_prompt_parse}
\end{figure*}

\begin{figure*}[!htbp]
\centering 
\includegraphics[width=\linewidth]{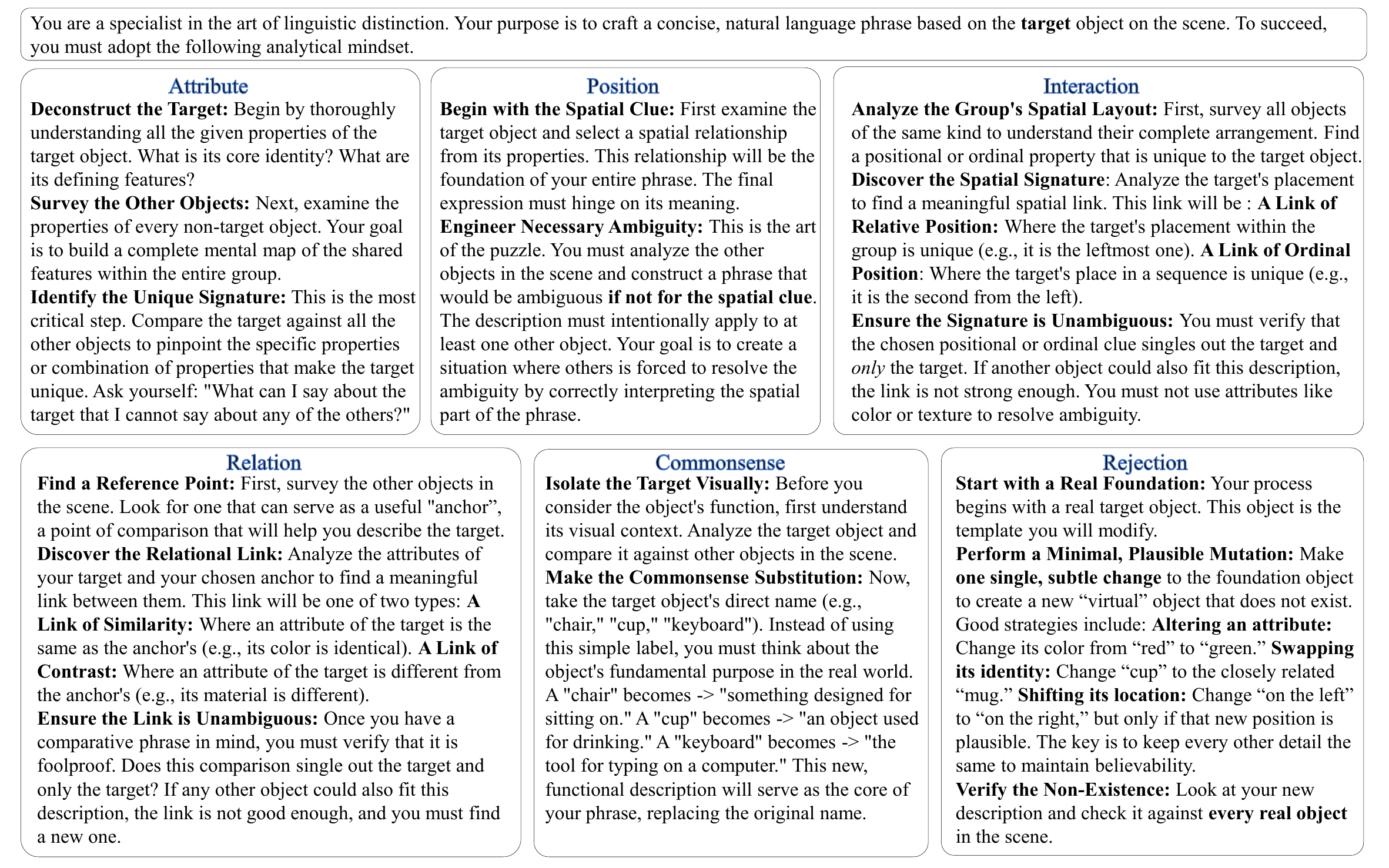}
\caption{Task-specific prompts for referring expression generation in PR-Bench.}
\label{fig:supp_prompt_expression}
\end{figure*}

\section{Additional Evaluation Details}
For COCO evaluation, we follow the T-Rex2~\cite{jiang2024t} protocol for all three settings. For MLLM-based methods, following Youtu-VL~\cite{wang2025x}, we use the predicted box area as the confidence score for mAP computation. We also report F1@mIoU as a complementary metric.

\section{Visualizations}
\subsection{Visualizations of PR-Bench}
We provide visualizations on PR-Bench in Figure~\ref{fig:supp_prbench_case}, covering all six subcategories, and showing the predictions of representative models to illustrate the complexity of PR-Bench.
\begin{figure*}[!htbp]
\centering 
\includegraphics[width=\linewidth]{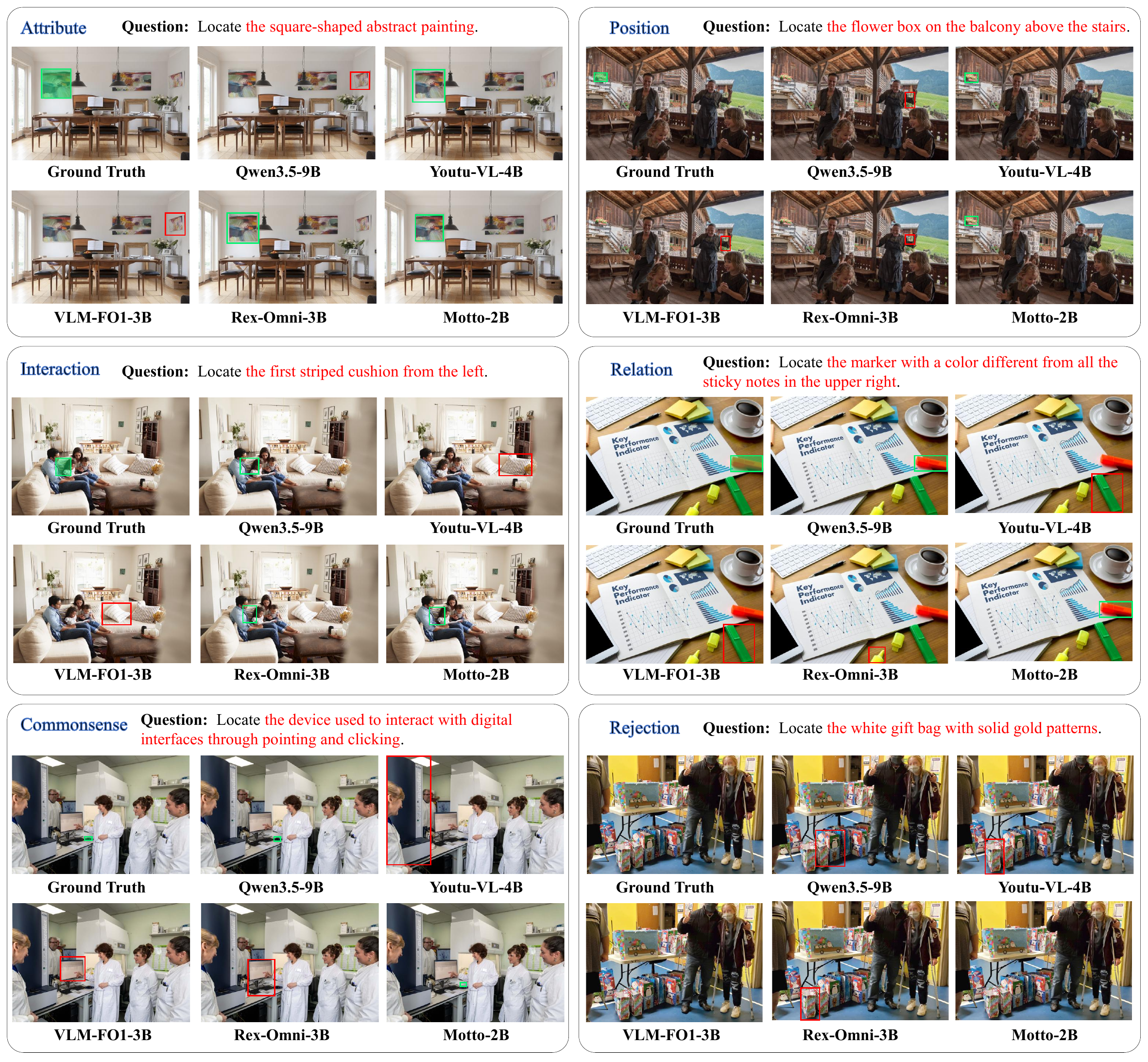}
\caption{Visualizations of cases on PR-Bench. Each row presents a challenging example and the corresponding predictions from representative models.}
\label{fig:supp_prbench_case}
\end{figure*}

\subsection{In-depth Case Study}
We provide an in-depth case study to further examine the behavior of Motto under different grounding scenarios. As shown in Figure~\ref{fig:supp_motto_case_study}, the proposed Context-Adaptive Chain-of-Tokens enables the model to dynamically adapt its grounding mode based on contextual semantics and the sufficiency of available evidence. 
When the evidence is sufficient, Motto directly decodes P-Tokens for grounding; otherwise, it switches to reasoning mode and incorporates textual reflection and visual evidence to support R-Token generation. These examples show how the Scan-Focus-Action paradigm enables adaptive grounding across different instances.

\begin{figure*}[!htbp]
\centering 
\includegraphics[width=\linewidth]{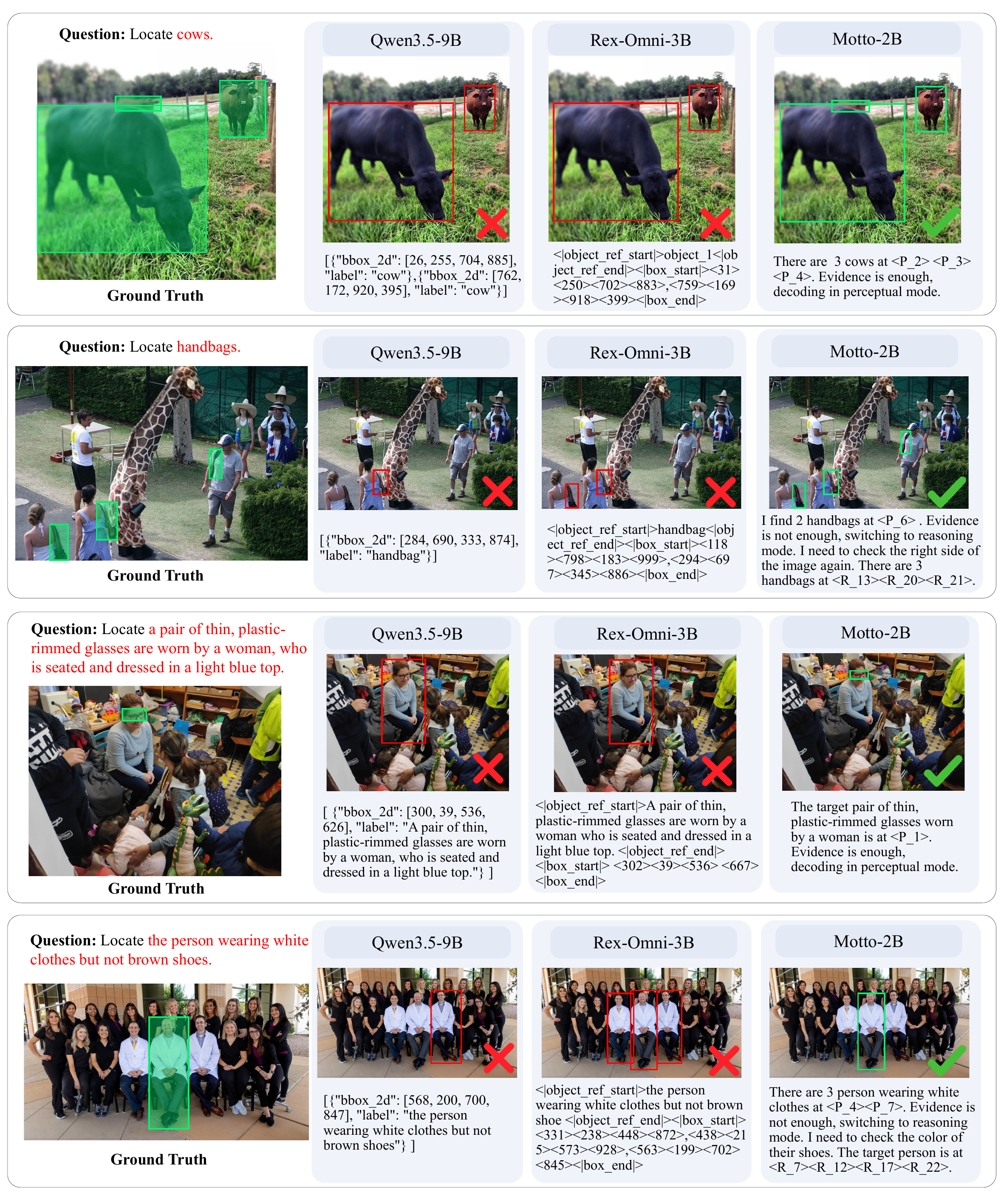}
\caption{In-depth case study of Motto under different grounding scenarios. Motto dynamically adapts its grounding mode through the Scan-Focus-Action paradigm, performing either direct perceptual grounding from P-Tokens or reasoning-driven grounding with re-focused semantic and visual evidence.}
\label{fig:supp_motto_case_study}
\end{figure*}

\subsection{Extended Visualization Results of Motto}
Figure~\ref{fig:supp_motto_visualize} presents additional visualization results across a wide range of grounding tasks, offering a more comprehensive and intuitive view of Motto's capabilities.

\begin{figure*}[!htbp]
\centering 
\includegraphics[width=\linewidth]{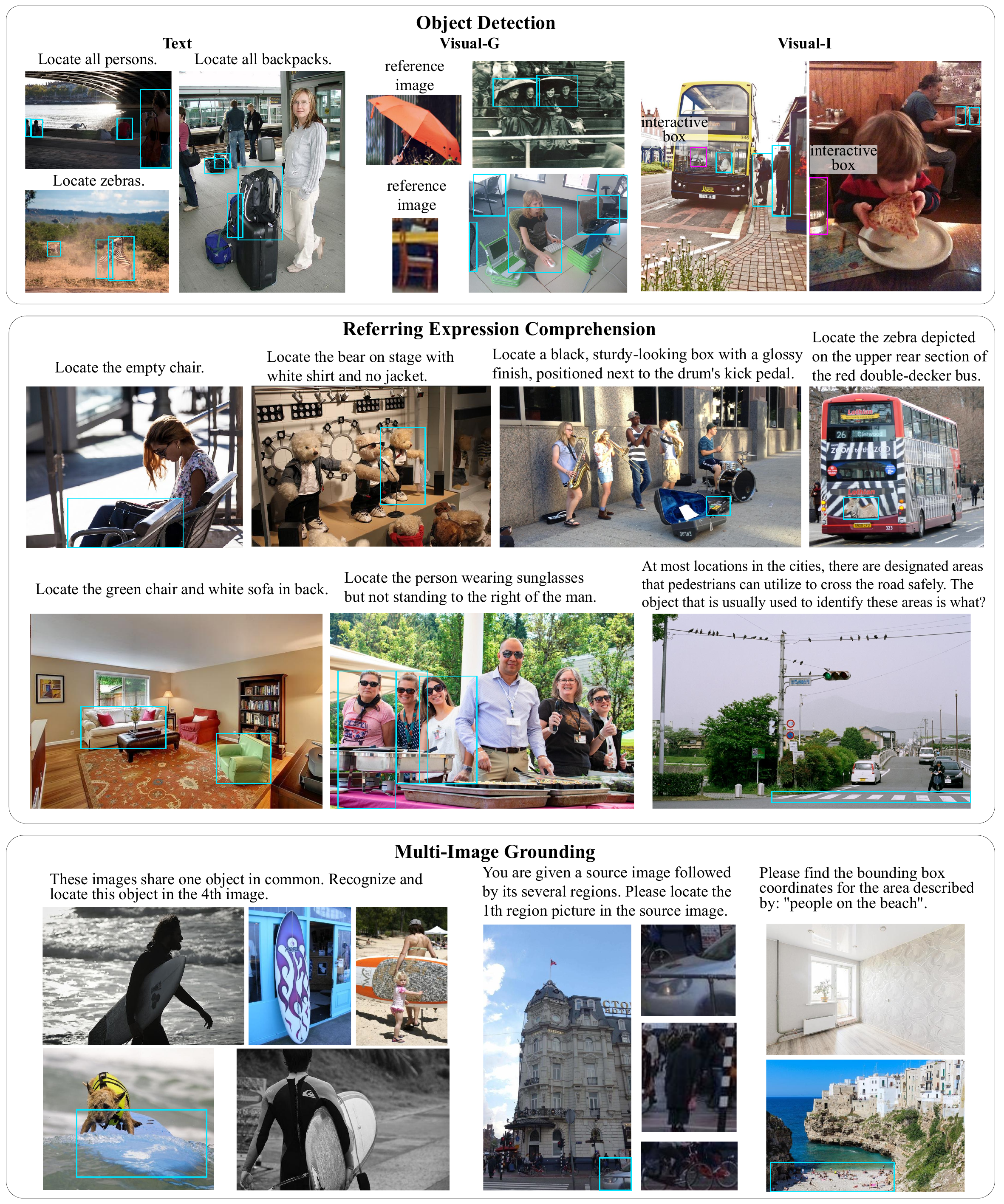}
\caption{Extended visualization results of Motto across diverse grounding tasks, including object detection under text, reference-image(Visual-G), and interactive visual (Visual-I) prompts, referring expression comprehension with inputs ranging from short phrases and long captions to reasoning questions, and targets from single to multiple objects, and multi-image grounding across multiple candidate images.}
\label{fig:supp_motto_visualize}
\end{figure*}

\end{document}